%% file: main.tex
\documentclass[runningheads]{llncs}

\usepackage{eccv}

\usepackage{xltabular}

\usepackage{eccvabbrv}

\usepackage{graphicx}
\usepackage{booktabs}

\usepackage[accsupp]{axessibility}  % Improves PDF readability for those with disabilities.

\usepackage{booktabs}
\usepackage{arydshln}
\usepackage{multirow}
\usepackage{makecell}
\usepackage{tabularx}

\usepackage{ulem}
\usepackage[inline]{enumitem}

\usepackage{soul}    % provides \hl and \st
\usepackage{xcolor}  % provides \textcolor
\usepackage{graphicx}
\usepackage{hyperref}

\usepackage{orcidlink}

\begin{document}

\makeatletter
\let\@oldmaketitle\@maketitle% Store \@maketitle
\renewcommand{\@maketitle}{\@oldmaketitle% Update \@maketitle to insert...
\input{figures/qualitative_results_v2}}
\makeatother

% \title{Zero-Shot Affordance Grounding via Large Vision-Language Models} 
\title{Token-Based Affordance Grounding with \\ Large Vision-Language Models}

\titlerunning{TokAG}

\author{Seung Il Lee\inst{1}\orcidlink{0009-0006-2069-2528} \and
Qinqian Lei\inst{2}\orcidlink{0009-0001-6429-2722} \and
Daguang Xu\inst{3}\orcidlink{0000-0002-4621-881X}
\and
Dong Yang\inst{3}\orcidlink{0000-0002-5031-4337} \\
Robby T. Tan\inst{2}\orcidlink{0000-0001-7532-6919}
\and
Yixin Chen\inst{1}\orcidlink{0009-0009-5275-3665}
\and
Bo Wang\inst{1}\orcidlink{0000-0002-0127-2281}
}

\authorrunning{Lee et al.}

\institute{$^1$University of Mississippi $\cdot$ $^2$National University of Singapore $\cdot$ $^3$NVIDIA
\\
% \email{drfirst.lee@gmail.com} \email{qinqian.lei@u.nus.edu} \email{hawk.rsrch@gmail.com}
% \\
\url{https://github.com/DrFirstLee/tokAG}
}
% \\ 
\maketitle

\begin{abstract}

% For Fig1, Delete GT and add Raw LVLM results with bbox and CLIPseg
% Our columns last column
% GT is little bit controversial delete GT and add VLM CLIP
% We have to add some explanation the way how we changed CLIPseg result into Heatmap

Affordance grounding aims to localize image regions that support a specific action, serving as a core capability for physical intelligence and embodied perception.  
Previous studies have primarily relied on weakly supervised learning with action labels from exocentric images.
However, these methods often struggle with visually ambiguous exocentric images containing co-occurring actions; moreover, they fail to distinguish semantically similar actions because existing methods typically rely on brief action phrases that lack rich semantic details for action-specific localization.
Although large vision-language models (LVLMs) encode rich action semantics and their action-conditioned textual outputs implicitly contain spatial cues, they do not directly provide action-specific spatial localization.
To address these problems, we propose \textbf{TokAG}, a zero-shot affordance grounding framework that exploits the token-level semantic-spatial signals in LVLMs to localize action-relevant regions without external supervision. 
We observe that attention maps associated with different LVLM output tokens vary significantly, with many attending to irrelevant regions such as the background.
Thus, we introduce a spatial-aware token-selection mechanism to systematically evaluate each output token and select the one whose attention maps exhibit dominant activation over the target object, instead of relying on arbitrary attention maps. 
By extracting these object-focused attention maps, we transform the LVLM’s implicit semantic signals into zero-shot affordance heatmaps.
Our zero-shot framework consistently outperforms prior weakly supervised approaches across multiple benchmarks, improving NSS by 10.7\% on the unseen split of AGD20K and by 29.7\% on HICO-IIF.
The code and models will be made publicly available. 

\end{abstract}

\section{Introduction}
\label{sec:intro}

The concept of affordance was introduced by Gibson~\cite{gibson2014ecological} to describe action possibilities offered by the environment. In computer vision, affordance grounding aims to localize specific image regions that support potential human actions~\cite{luo2022learning}.
This task goes beyond object recognition, serving as a core capability for physical intelligence and embodied perception. Accurately understanding actionable regions is crucial for empowering a wide range of downstream applications, including robotic manipulation, human-computer interaction, and augmented reality, enabling artificial agents to navigate and interact with the 3D physical world safely~\cite{ardon2020affordances,hassanin2021visual,ma2024local,bahl2023affordances,brohan2023can,yang2023grounding}.  

Despite recent progress, accurately grounding affordances remains challenging, especially in complex real-world scenarios where action understanding and spatial localization must work together.
Previous studies primarily approach affordance grounding through weakly supervised learning, leveraging exocentric images with action-level supervision, where actions are typically represented as brief phrases.
However, these methods~\cite{li2023locate,jang2024intra,tang2025closed} struggle with visually ambiguous exocentric images containing co-occurring actions. Our analysis shows that 31.7\% of exocentric images are reused across different classes, capturing multiple actions (e.g., ``brush with'' and ``hold'' toothbrush) simultaneously. Such ambiguity prevents models from isolating action-specific visual evidence, leading to mislocalized predictions. 
As illustrated in the bottom two rows of \cref{fig:representative_result}, existing methods often produce identical heatmaps regardless of whether the action is ``cut with'' or ``hold.''

Moreover, existing methods~\cite{xu2024weakly,moon2025selective,xu2025weaklysupervised} fail to distinguish semantically similar actions on the same object. This limitation arises because they rely on brief action phrases that lack the semantic detail needed to disambiguate fine-grained actions and localize action-specific regions \cite{lei2025hola}.
As shown in the first and second rows of \cref{fig:representative_result}, existing methods struggle to differentiate nuanced actions such as ``push'' and ``sit on'' on the same motorcycle.
While both actions involve the same object, they require grounding in distinct parts (the seat for sitting on and the handle for pushing), which existing methods often fail to isolate using weak action phrases alone.

Recently, large vision-language models (LVLMs) have demonstrated promising capabilities in visual understanding~\cite{bai2025qwen3,zhu2025internvl3,wu2024deepseek, cao2026smart,liu2026tele, lei2024ez}. 
These models encode rich action semantics, and their action-conditioned textual outputs implicitly contain spatial cues grounded in the visual input.
However, LVLMs are primarily optimized for text generation rather than fine-grained, action-specific spatial localization \cite{lei2026crosshoi_bench}. Their internal attention mechanisms are not explicitly designed to produce precise affordance maps, making direct extraction of spatially accurate regions non-trivial. 

Motivated by these observations, we propose \textbf{TokAG (Token-based Affordance Grounding)}, a zero-shot affordance grounding framework that leverages the cross-modal alignment between vision and language representations within LVLMs. 
Our approach directly localizes action-relevant regions without requiring external annotations or image-level supervision, enabling affordance grounding in a zero-shot setting. Importantly, our framework does not involve task-specific training or supervision for affordance localization.
We observe that spatial activations within LVLMs vary substantially across output tokens, architectural layers, and attention heads, often exhibiting diffuse patterns that respond to irrelevant background context rather than target functional regions. This variability makes naive extraction of attention maps unreliable for affordance localization.
Instead of relying on structural pruning~\cite{michel2019sixteen} or layer and head selection strategies~\cite{kang2025your} that require supervision, we leverage the full cross-attention hierarchy of the LVLM and guide it using object-level spatial constraints. By aggregating activations across layers and heads, we preserve rich semantic-spatial cues while suppressing irrelevant background responses.
Within this refined attention field, we identify the output token whose spatial activation is most concentrated on the target region. By isolating this token-specific activation map, we transform the LVLM’s implicit semantic knowledge into precise affordance heatmaps without any task-specific supervision.
In summary, our main contributions are as follows:
\begin{itemize}
    \item We introduce \textbf{TokAG}, a zero-shot affordance grounding framework that leverages latent semantic-spatial signals within pretrained LVLMs, eliminating the need for affordance-specific supervision.

    \item We propose a spatial-aware token-selection mechanism that aggregates cross-attention activations across layers and heads, and identifies object-focused tokens to produce precise affordance heatmaps.

    \item We provide an empirical analysis of the tokens selected by our framework, showing that they often correspond to action or object cues, and stronger LVLM backbones tend to select more semantically relevant tokens, which correlates with improved affordance grounding performance.
    
\end{itemize}

% \hawk{Thoughts for revising the contribution list:
% If We can show: The selected token is semantically meaningful.
% It can help us:
% 1. Strengthens the contribution. From "token selection heuristic" to "interpretable semantic token grounding mechanism."
% 2. Answer possible questions from the reviewers (Why should this token represent the affordance?) Our answer: Empirically the selected tokens correspond to meaningful affordance-related semantics. 
% }

% \hawk{We also want to be careful to not overclaim interpretability. We may consider to use these phrases: interpretable token, semantically meaningful token, action-relevant token. Avoid using: explainable AI, interpretability framework, explainability method, because the reviewers might be sensitive to these phrases.}

% \hawk{If the experiment works, our contributions could become: 
% 1. Zero-shot affordance grounding via LVLM semantic signals. 2. Semantic token discovery mechanism for affordance localization. 3. Semantic-guided refinement for spatial grounding.} 

Notably, our zero-shot framework achieves an NSS of 1.514 on the unseen split of AGD20K, demonstrating a 10.7\% improvement over the prior state-of-the-art (1.368). Furthermore, on the HICO-IIF dataset, our method reaches an NSS of 1.655, representing a 29.7\% increase compared to existing weakly supervised baselines (1.234).

\section{Related Work}
\label{sec:related_work}

\subsection{Weakly Supervised Affordance Grounding}

Affordance grounding aims to localize object regions that support specific
human actions~\cite{gibson2014ecological}. Due to the high cost of obtaining pixel-level affordance, earlier works~\cite{chuang2018learning,
koppula2013learning, myers2015affordance} primarily relied on fully supervised
training, limiting scalability in complex real-world environments.
To alleviate this issue, recent studies~\cite{sawatzky2017weakly,
mai2020erasing, pan2021unveiling, gao2021ts} adopt weakly supervised learning
paradigms that require only action-level supervision. These approaches can be
broadly categorized into two groups based on their source of knowledge
transfer.

\par\vspace{0.5em}

\noindent \textbf{Vision-Only Knowledge Transfer.}
The first line of research focuses on transferring visual knowledge from exocentric images with action-level supervision. Cross-View-AG~\cite{luo2022learning} first
introduced this paradigm by adopting the Class Activation Map
(CAM)~\cite{zhou2016learning} mechanism to extract affordance cues from exocentric views in a weakly supervised manner. 
Subsequent methods have further improved this framework through more sophisticated transfer mechanisms.
LOCATE~\cite{li2023locate} groups interaction embeddings from exocentric images into compact prototypes representing humans, object parts, and background regions to guide affordance grounding. 
INTRA~\cite{jang2024intra} learns interaction representations directly from exocentric images, removing the need for paired datasets, while capturing unique interaction features through a contrastive learning framework.
More recently, LoopTrans~\cite{tang2025closed} introduces a closed-loop framework that enables bidirectional knowledge transfer between exocentric and egocentric views through a denoising knowledge distillation process.
However, these approaches often struggle with visually ambiguous
exocentric images containing multiple co-occurring actions, making it difficult to isolate action-specific affordance cues.

\par\vspace{0.5em}

\noindent \textbf{Vision-Language Knowledge Transfer.}
The second group incorporates textual information to improve affordance localization. 
WSMA~\cite{xu2024weakly} leverages both exocentric images and
textual action labels to guide affordance localization in egocentric views through an HOI-Transfer module and a Pixel-Text Fusion module. It also introduces learnable prompts to refine the semantic representation of affordances.
WSAG-PLSP~\cite{xu2025weaklysupervised} further explores this
direction by employing large language models to refine affordance part descriptions and generate pseudo-labels using vision models such as VLPart~\cite{sun2023going} and SAM~\cite{kirillov2023segment}. These pseudo-labels provide additional supervision during training. 
Meanwhile, Selective Contrastive Learning~\cite{moon2025selective} utilizes 
CLIP~\cite{radford2021learning} to transfer text-conditioned affordance knowledge across exocentric and egocentric views through selective contrastive objectives.
Despite these advances, existing methods still rely on short action phrases, which often lack the rich semantic detail needed to distinguish semantically similar actions and accurately localize action-specific regions.

\subsection{Visual Understanding and Localization in LVLMs}

\noindent \textbf{Large Vision-Language Models (LVLMs).}
Recent large vision-language models (LVLMs), such as Qwen3-VL~\cite{bai2025qwen3} and InternVL3~\cite{zhu2025internvl3}, have demonstrated promising progress in open-vocabulary visual understanding. 
These models achieve strong zero-shot performance on challenging multimodal benchmarks such as MMMU~\cite{yue2024mmmu} and MathVista~\cite{lu2023mathvista}, which require reasoning beyond basic object recognition. 
Their ability to interpret complex visual scenes and generate descriptive textual outputs, such as identifying rare objects or explaining relationships within an image, highlights their
strong semantic reasoning  capabilities~\cite{bai2025qwen3,zhu2025internvl3,wu2024deepseek}.

\par\vspace{0.5em}

\noindent \textbf{Localization Efforts in LVLMs.}
Recent studies have explored training-free approaches for visual grounding by analyzing the internal states of multimodal large language models.
Kang et al.~\cite{kang2025your} show that visual grounding can be achieved by leveraging a small subset of attention heads guided by the final token of the input prompt, while also identifying visual attention sinks that must be suppressed to improve localization accuracy~\cite{kang2025see}. 
Zhang et al.~\cite{zhang2025mllms} further observe that LVLMs may fail to explicitly describe small visual details. However, their internal attention patterns often correctly indicate where the model focuses its attention within the image. Based on this observation, they propose a training-free intervention strategy that uses attention and gradient maps to identify and crop relevant regions, improving the perception of small objects during inference. 
These studies suggest that the cross-modal attention mechanisms of LVLMs encode useful spatial cues for visual localization. However, existing approaches utilize these attention signals to enhance text generation capabilities, or they rely on the heuristic selection of attention heads or tokens, which may not reliably isolate action-specific regions required for affordance grounding.

\begin{figure}[t!]
    \centering
    \includegraphics[width=\linewidth]{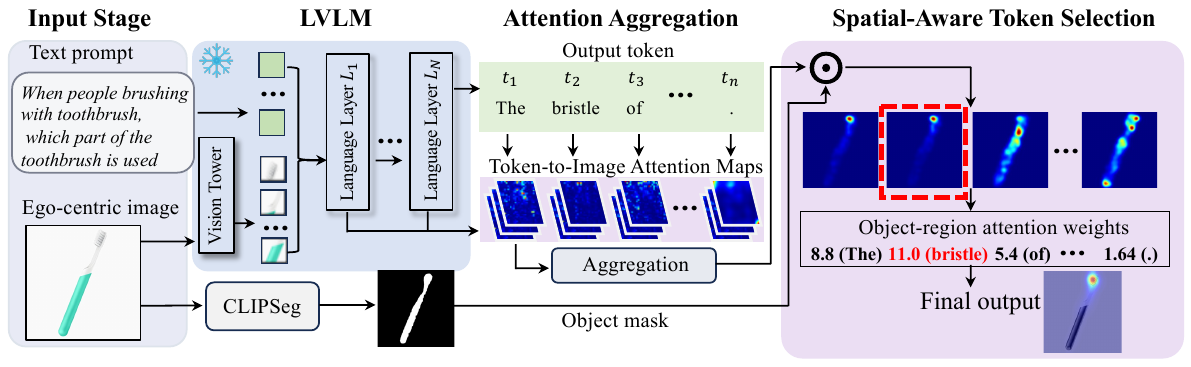}
    \caption{Overview of the \textbf{TokAG} framework. Given an egocentric  image and a text prompt, we extract output tokens (e.g., $t_1, t_2, t_3$) from the language layers of a frozen LVLM. We aggregate token-to-image attention maps across all layers and heads. With object masks generated by CLIPSeg~\cite{luddecke2022image}, we compute object-region attention weights for each token. The token exhibiting the highest score is selected (e.g., $t_2$, corresponding to ``bristle'' with a score of 11.0). The object-region attention heatmap corresponding to the selected token becomes the final prediction.}
    \label{fig:pipeline}
\end{figure}

\section{Method}
\label{sec:method}

We propose a zero-shot affordance grounding framework that localizes action-specific regions by extracting the implicit spatial knowledge of large vision-language models (LVLMs) without relying on external supervision. 
As illustrated in Fig.~\ref{fig:pipeline}, our pipeline takes an egocentric image and a plain text prompt asking about the affordance area on a target object as inputs. 
Our method extracts and aggregates cross-modal attention maps utilizing a spatial-aware token selection mechanism to identify the most relevant action-conditioned spatial cues (Section~\ref{subsec:token_selection}). 
Furthermore, we observe that the selected tokens can often be decoded into meaningful semantic phrases of functional object parts, making our token selection mechanism interpretable (Section~\ref{subsec:token_analysis}).

\subsection{Attention Aggregation and Spatial-Aware Token Selection}
\label{subsec:token_selection}
\begin{figure}[t]
    \centering
    \includegraphics[width=\linewidth]{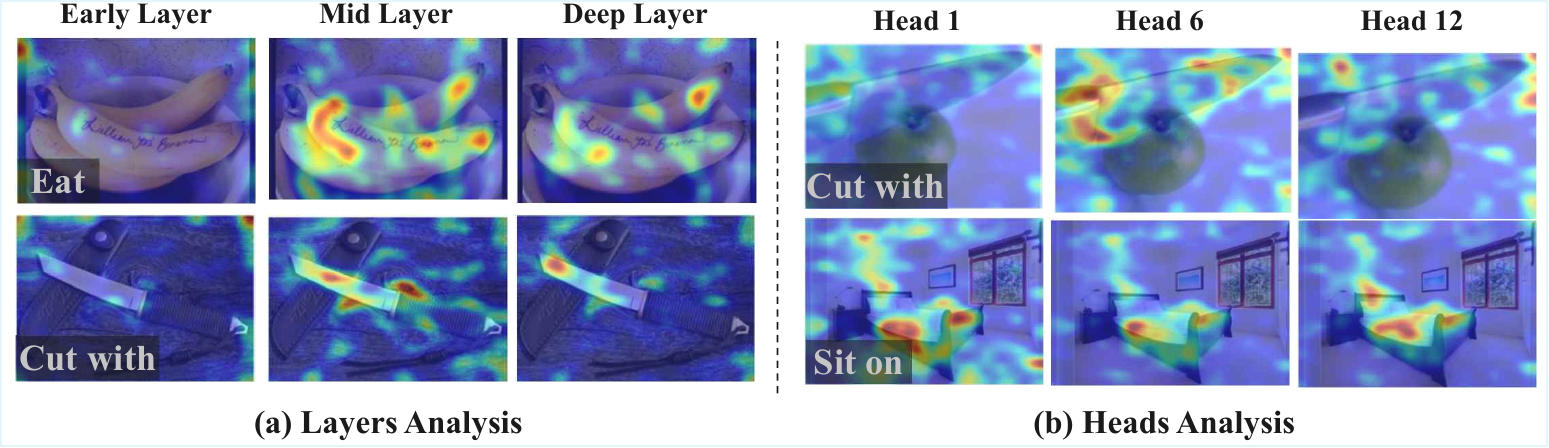}
    \caption{Attention maps across various layers and heads. (a) Attention maps extracted from early, middle, and deep layers show different spatial activation patterns. (b) Different attention heads also focus on distinct image regions. These observations indicate that affordance-relevant cues are distributed across layers and heads, while some activations attend to task-irrelevant background regions.
    }
    \label{fig:attn_vary}
\end{figure}

Given an input image $I$, a target object $O$, and an action $A$, we formulate a simple text prompt $P$ to query the LVLM: \textit{``When people perform $A$ with $O$, which part of the $O$ is used for `$A$'? Answer in one sentence.''} Based on these input tokens, the LVLM generates a sequence of output tokens $T = \{t_1, t_2, \dots, t_N\}$. 
While LVLMs inherently encode rich action semantics, their textual responses do not provide explicit spatial coordinates. 
To bridge this gap, we analyze the cross-modal attention maps between generated output tokens and visual patches.
As illustrated in Fig.~\ref{fig:attn_vary}, two key observations emerge.
First, the spatial activations across layers and heads are diverse, often focusing on different image regions.
Second, many attention responses are diffuse and may attend to task-irrelevant background regions due to the autoregressive generation process~\cite{kang2025see}.
Meanwhile, we also observe that although the attention distributions vary across layers and heads, the activations that fall within the object region are typically related to the functional parts involved in the queried action.
For example, when querying the action ``sit on bed'', the model consistently highlights the mattress region.
This behavior is intuitive because the attention maps originate from action-conditioned output tokens generated by the LVLM.

These observations motivate our spatial-aware token selection design.
Instead of selecting specific layers or heads, we aggregate attention weights across all layers and attention heads to capture the distributed semantic–spatial cues within the LVLM. 
Formally, for each output token $t_i$, we obtain an aggregated attention map
$\mathcal{M}_{t_i} \in \mathbb{R}^{H \times W}$ by integrating attention responses across the architectural layers and attention heads.
Moreover, since attention activation inside the object region indicates useful affordance cues, we introduce an object-aware spatial constraint to suppress background noise and emphasize attention responses within the target object.
To achieve this, we leverage a binary object mask $M_{obj} \in \{0,1\}^{H \times W}$ for the target object $O$, generated by CLIPSeg~\cite{luddecke2022image}.

First, we define the masked attention map $\widetilde{\mathcal{M}}_{t_i}$ by applying the object mask to the aggregated attention map via element-wise multiplication:
\begin{equation}
    \widetilde{\mathcal{M}}_{t_i} = \mathcal{M}_{t_i} \odot M_{obj}
\end{equation} where $\odot$ denotes the Hadamard product.
We then compute the overlap score $S_i$ by aggregating the activation values of the masked attention map:
\begin{equation}
    S_i = \sum_{x,y} \widetilde{\mathcal{M}}_{t_i}(x,y)
\end{equation}
Then, we select the single output token $t^*$ that yields the highest activation score over the target object area:
\begin{equation}
    t^* = \underset{t_i \in T}{\mathrm{argmax}} \, S_i
\end{equation}
Finally, the heatmap $\mathcal{H}$ is derived by upsampling the selected masked attention map $\widetilde{\mathcal{M}}_{t^*}$ to the original image dimensions using bilinear interpolation: $\mathcal{H} = \mathrm{UP}(\widetilde{\mathcal{M}}_{t^*})$.

\subsection{Analysis of Selected Tokens}
\label{subsec:token_analysis}

\begin{figure}[t]
    \centering
    \includegraphics[width=\linewidth]{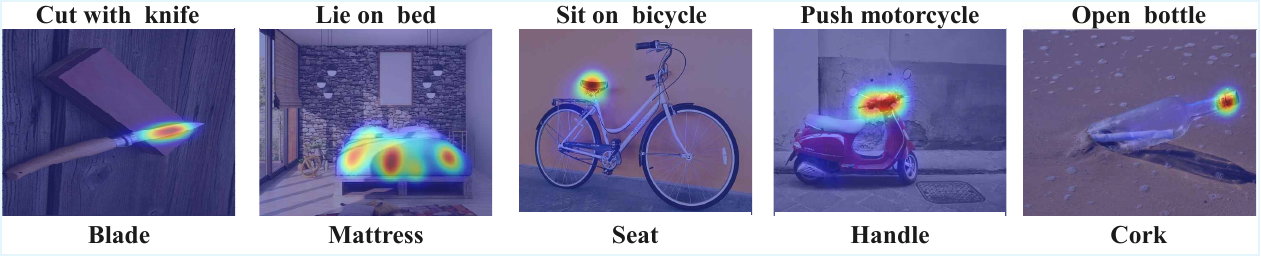}
    \caption{Examples of selected tokens. For each object–action pair, our spatial-aware token selection mechanism identifies the output token whose attention is most concentrated on the target object region. The selected token is then decoded into text for visualization. Interestingly, many decoded tokens correspond to meaningful object parts or functional regions (e.g.,  ``blade'', ``mattress'', ``seat'', ``handle'', and ``cork''), which naturally align with the affordance-relevant areas.}
    \label{fig:interpret_tok}
    % \vspace{-5mm}
\end{figure}

Fig.~\ref{fig:interpret_tok} presents qualitative examples of the decoded texts corresponding to the output tokens selected by our spatial-aware token selection mechanism.
Interestingly, many of these tokens correspond to meaningful object parts or functional regions, such as ``blade'', ``mattress'', ``seat'', ``handle'', and ``cork''. These tokens naturally describe the regions that support the queried actions, suggesting that the LVLM implicitly encodes functional knowledge about objects.

To further analyze this phenomenon, we perform a semantic categorization of the selected tokens across the dataset, as summarized in Table~\ref{tab:semantic_analysis}. 
For the 2B model, approximately 55\% of the selected tokens correspond to
semantically meaningful action- or object-related words. This
percentage increases to 65\% when using the 32B model, suggesting that
stronger semantic representations lead to more reliable token selection. This trend aligns with the improved affordance grounding performance
observed with larger LVLM backbones.

The key message is not that all the selected tokens are always meaningful, but that better models tend to select more meaningful tokens, and this correlates with better affordance localization. This gives us a more interpretable explanation of why the token-selection mechanism works and also suggests a possible direction for improving affordance grounding in the future.

\begin{table}[t]
\centering
\caption{\textbf{Semantic Analysis of Top-1 Tokens.} Empirical results show that, the selected tokens correspond to meaningful affordance-related semantics. Specifically, 54.3\% of the Top-1 tokens in the 2B model and 62.5\% in the 32B model were identified as meaningful, indicating stronger semantic grounding with larger model sizes. The categories were initially assigned using GPT-based labeling and then verified by a human reviewer. The semantic categories are defined as follows: Meaningful Action (verbs such as `sit' or `hold'), Object-Related Area (spatial terms like `top', `bottom', or `edge'), Function-Related (functional parts like `handle', `seat', or `blade'), and Others (grammatical articles like `a' or `the').}
\label{tab:semantic_analysis}
\begin{tabular}{l|c |c |c | c}
\toprule
{Model} & {Meaningful Action} & {Object-Related Area} & {Function-Related} & {Others} \\
\midrule
2B  & 23.0\% & 9.8\%  & 21.5\% & 45.7\% \\
32B & 1.9\%  & 20.2\% & 40.4\% & 37.6\% \\
\bottomrule
\end{tabular}
\end{table}

\begin{table*}[t]
\vspace{-2mm}
\centering
\caption{Performance comparison on the AGD20K dataset. Set 1 is the Seen test set, Set 2 is the Unseen test set.
Lower is better for KLD ($\downarrow$), higher is better for SIM/NSS ($\uparrow$).
The best performance is \textbf{bolded} and the second-best is \underline{underlined}.}
\label{tab:agd20k_main}
\vspace{-2mm}
\renewcommand{\arraystretch}{1.1}
\begin{tabular*}{\textwidth}{@{\extracolsep{\fill}} l ccc ccc @{}}
\toprule
\multirow{2}{*}{Method} &
\multicolumn{3}{c}{AGD20K-Set 1} &
\multicolumn{3}{c}{AGD20K-Set 2} \\
\cmidrule(lr){2-4} \cmidrule(lr){5-7}
& KLD$\downarrow$ & SIM$\uparrow$ & NSS$\uparrow$
& KLD$\downarrow$ & SIM$\uparrow$ & NSS$\uparrow$ \\
\midrule

Cross-View-AG (CVPR'22)~\cite{luo2022learning}
& 1.538 & 0.334 & 0.927
& 1.787 & 0.285 & 0.829 \\

Cross-View-AG+ (CVPR'22)~\cite{luo2022learning}
& 1.489 & 0.342 & 0.981
& 1.765 & 0.279 & 0.882 \\

LOCATE (CVPR'23)~\cite{li2023locate}
& 1.226 & 0.401 & 1.177
& 1.405 & 0.372 & 1.157 \\

INTRA (ECCV'24)~\cite{jang2024intra}
& 1.199 & 0.407 & 1.239
& 1.365 & 0.375 & 1.209 \\

WSMA (AAAI'24)~\cite{xu2024weakly}
& 1.176 & 0.416 & 1.247
& 1.335 & 0.382 & 1.220 \\

LoopTrans (ICCV'25)~\cite{tang2025closed}
& 1.088 & 0.445 & 1.322
& 1.247 & 0.403 & 1.315 \\

Moon et al. (ICCV'25)~\cite{moon2025selective}
& 1.124 & 0.433 & 1.280
& 1.243 & 0.405 & 1.368 \\

\midrule
\textbf{Ours} (2B)
& \underline{1.020} & \textbf{0.459} & \underline{1.406} 
& \underline{1.060} & \textbf{0.455} & \underline{1.514}\\

\textbf{Ours} (32B)
& \textbf{1.010} & \underline{0.458} & \textbf{1.424} 
& \textbf{1.043} & \textbf{0.455} & \textbf{1.549}\\
\bottomrule
\end{tabular*}
% \vspace{-2mm}
\end{table*}

\begin{table}[t]
% \vspace{-2mm}
\centering
\caption{Performance comparison on the HICO-IIF dataset (Seen test set).
Lower is better for KLD ($\downarrow$), higher is better for SIM/NSS ($\uparrow$).
The best performance is \textbf{bolded} and the second-best is \underline{underlined}.}
\label{tab:hico_iif}
\vspace{-2mm}
\setlength{\tabcolsep}{7pt}
\renewcommand{\arraystretch}{1.1}
\begin{tabular}{l ccc}
\toprule
Method & KLD$\downarrow$ & SIM$\uparrow$ & NSS$\uparrow$ \\
\midrule

Cross-View-AG+ (CVPR'22)~\cite{luo2022learning}
& 1.779 & 0.263 & 0.946 \\

LOCATE (CVPR'23)~\cite{li2023locate}
& 1.593 & 0.327 & 0.966 \\

WSMA (AAAI'24)~\cite{xu2024weakly}
& 1.465 & 0.358 & 1.012 \\

LoopTrans (ICCV'25)~\cite{tang2025closed}
& 1.399 & 0.379 & 1.226 \\

Moon et al. (ICCV'25)~\cite{moon2025selective}
& 1.358 & 0.378 & 1.234 \\

\midrule
\textbf{Ours} (2B)
& \underline{1.056} & \textbf{0.459} & \underline{1.600} \\

\textbf{Ours} (32B)
& \textbf{1.032} & \underline{0.449} & \textbf{1.655} \\
\bottomrule
\end{tabular}
\vspace{-2mm}
\end{table}

\section{Experiments}
\label{sec:exp}

\subsection{Experimental Setup}

\textbf{Datasets.} 
We evaluate our method on two benchmark datasets for affordance grounding: AGD20K~\cite{luo2022learning} and  
HICO-IIF~\cite{xu2024weakly}. 
AGD20K is a large-scale affordance grounding benchmark containing 
20,061 exocentric images and 3,755 egocentric images annotated with 
36 affordance categories. The dataset defines two evaluation settings 
based on action-object pair distributions. In the \textit{Seen} setting, the training and test splits share identical 
action-object pairs (36 actions and 50 objects), differing only in the 
egocentric images used for evaluation. In the \textit{Unseen} setting, 
the train and test splits share the same action categories but involve 
non-overlapping object sets (e.g., training on ``hold badminton racket'' 
and testing on ``hold axe''), which evaluates the model’s ability to 
generalize affordance understanding across different objects.

Existing weakly supervised methods~\cite{luo2022learning,li2023locate,
jang2024intra,xu2024weakly,tang2025closed,moon2025selective,
xu2025weaklysupervised} rely on transferring knowledge from exocentric 
training images and therefore explicitly train and evaluate under the 
\textit{Seen} and \textit{Unseen} settings. In contrast, our method operates 
in a zero-shot manner without using any training data from AGD20K. 
To avoid confusion with training-based protocols, we treat the two 
conventional evaluation settings as independent test splits, which we 
refer to as \textbf{Set 1} (Seen) and \textbf{Set 2} (Unseen). Apart from AGD20K dataset, HICO-IIF~\cite{xu2024weakly} is an affordance grounding benchmark derived from the HICO-DET dataset~\cite{chao2018learning}, containing human-object interaction images with pixel-level affordance annotations. Following existing methods~\cite{xu2024weakly,tang2025closed,moon2025selective}, we evaluate our method on this dataset to assess generalization across different interaction scenarios and datasets.
\par\vspace{0.5em}
\noindent \textbf{Evaluation Metrics.} Following standard evaluation protocols used in existing methods, we adopt three widely-used evaluation metrics to quantify the alignment between the predicted affordance heatmaps and the ground truth maps: Kullback-Leibler Divergence ($KLD$), Similarity ($SIM$), and Normalized Scanpath Saliency ($NSS$). Higher values for $SIM$ and $NSS$, and lower values for $KLD$, indicate better grounding performance.
\par\vspace{0.5em}
\noindent \textbf{Implementation Details.} Our method utilizes Qwen3-VL~\cite{bai2025qwen3} as the LVLM to generate descriptive tokens and extract internal multi-head attention maps. 
For the spatial-aware token selection and the subsequent object-centric spatial alignment, we use the pre-trained CLIPSeg~\cite{luddecke2022image} model. 
All input images are uniformly resized to a spatial resolution of $1000 \times 1000$ pixels prior to processing.
All experiments, including the LVLM inference and token-centric attention extraction, were  conducted using an NVIDIA RTX 3090 GPU for the 2B model (an NVIDIA A100 GPU is used for the 32B model).

\subsection{Comparison to State-of-the-art Methods}
\label{subsec:major_comparison}

We compare our zero-shot framework against existing weakly-supervised affordance grounding methods~\cite{luo2022learning, li2023locate, jang2024intra, xu2024weakly, tang2025closed, moon2025selective} on the AGD20K dataset (Table~\ref{tab:agd20k_main}). Unlike these baselines, which rely on task-specific training on exocentric images, our approach operates in a zero-shot manner without using any training data.
Our method demonstrates strong generalization to novel object categories (Set~2). As shown in Table~\ref{tab:agd20k_main}, the \textbf{Ours} (2B) model achieves the lowest KLD (1.060) and the highest SIM (0.455) and NSS (1.514), outperforming previous best methods such as LoopTrans~\cite{tang2025closed} (KLD: 1.247) and Moon et al.~\cite{moon2025selective} (NSS: 1.368). These improvements highlight the ability of our framework to translate the implicit semantic knowledge of LVLMs into zero-shot affordance heatmaps.

Even on seen objects (Set~1), where baselines benefit from being trained on identical action–object pairs, our method remains competitive and achieves improved performance across all metrics. Specifically, the \textbf{Ours} (2B) model obtains the lowest KLD (1.020) and the highest SIM (0.459) and NSS (1.406). Finally, experiments with a larger backbone (32B) further improve performance across most metrics.
To further evaluate the robustness and generalization ability of our framework, we also conduct experiments on the HICO-IIF dataset~\cite{xu2024weakly}. As summarized in Table~\ref{tab:hico_iif}, our method consistently outperforms all weakly-supervised baselines. In particular, the \textbf{Ours} (2B) model achieves the highest SIM (0.459) and NSS (1.600), while the larger \textbf{Ours} (32B) model further improves performance with the lowest KLD (1.032) and the highest NSS (1.655). These results demonstrate that our zero-shot framework generalizes well across datasets and interaction scenarios.

\noindent \textbf{Runtime Evaluation.} While TokAG delivers superior zero-shot performance, it entails higher computational cost due to the autoregressive inference of LVLM backbones. Specifically, TokAG operates at 0.59 FPS (2B) and 0.20 FPS (32B), whereas the weakly-supervised baseline WSMA~\cite{xu2024weakly} runs at 20.4 FPS. However, this inference overhead is reasonable because TokAG is training-free and does not require task-specific optimization.

\noindent \textbf{Evaluation on Additional Dataset.} 
We evaluate TokAG on EPIC-Aff\cite{mur2023multi}, a challenging benchmark derived from EPIC-Kitchen\cite{damen2020epic}. We randomly sampled 500 images under the unseen condition, matching the scale of the AGD20K test set (540 images). On this dataset, TokAG outperforms WSMA across all metrics, achieving a lower KLD (3.367 vs. 3.372) and higher SIM (0.122 vs. 0.062) and NSS (0.927 vs. 0.413). As EPIC-Aff differs from object-centric affordance grounding benchmarks like AGD20K. EPIC-Aff contains less object-centric GT regions, which explains why the numbers are relatively low.

\subsection{Evaluation of Foundation-Model-based Baselines}

\label{subsec:ablation}

Although recent LVLMs demonstrate promising visual understanding capabilities, their standard textual outputs do not provide explicit spatial localization. Alternatively, vision foundation models (VFMs) such as SAM 3 \cite{carion2025sam3segmentconcepts} offer promising segmentation capabilities.
% \hawk{In the above place, the issue is similar to the caption of Table 4, we may check which term the reviewer used, vision foundation models is a bit narrow as we use LLM and SAM, LLM is not vision foundation model. Foundation-model-based baselines is the term the reviewers used, it is more suitable to be used to refer to Method E and F.}
%
To evaluate these alternative baselines with zero-shot capabilities, we compare TokAG with two groups of foundation-model-based baselines: (i) direct LVLM prompting methods for spatial coordinates, and (ii) multi-stage pipelines combining foundation models (e.g., LVLM + SAM 3 or GroundingDINO \cite{liu2024grounding} + SAM 3).
The results are summarized in Table~\ref{tab:lvlm_comparison_unseen}. 

\begin{table}[t]
\centering
\small 
\caption{Comparison with direct LVLM prompting and combined foundation-model baselines on the AGD20K Unseen (Set 2) dataset. (A) LVLM textual response; (B) Bounding box prediction from LVLM; (C) Gaussian-blurred anchor points from LVLM coordinate predictions; (D) Hadamard product of the predicted bounding box from LVLM and CLIPSeg mask; (E) Method B for bounding box prediction followed by segmentation with SAM 3; (F)
Extracting object bounding boxes via Grounding-DINO followed by segmentation with SAM 3.}

\label{tab:lvlm_comparison_unseen}
\setlength{\tabcolsep}{15pt} 
\renewcommand{\arraystretch}{1.2}
\begin{tabular}{c l c c c}
\toprule
\textbf{ID} & \textbf{Method} & \textbf{KLD $\downarrow$} & \textbf{SIM $\uparrow$} & \textbf{NSS $\uparrow$} \\
\midrule
A & LVLM\textsuperscript{1} & \multicolumn{3}{c}{Not available (Output was plain text)} \\
B & LVLM Bbox\textsuperscript{2} & 5.336 & 0.347 & 1.085 \\
C & LVLM Coord.\textsuperscript{3} & 9.054 & 0.170 & 0.675 \\ 
\midrule
D & B + CLIPseg\textsuperscript{4} & 5.684 & 0.346 & 1.085 \\
E & B + SAM 3 & 3.229 & 0.438 & 1.189 \\
F & \cite{liu2024grounding} + SAM 3 & 3.110 & 0.319 & 0.657 \\ 
\midrule
G & \textbf{TokAG (Ours)} & \textbf{1.060} & \textbf{0.455} & \textbf{1.514} \\
\bottomrule 
\noalign{\smallskip}
\multicolumn{5}{p{0.85\textwidth}}{\textsuperscript{1}\small Prompt: ``Which part of [object] is used for [action]?''} \\
\multicolumn{5}{p{0.85\textwidth}}{\textsuperscript{2}\small Add to prompt: ``Output the bounding box of this part in [xmin, ymin, xmax, ymax] format. Scale coordinates from 0 to 1000.''} \\
\multicolumn{5}{p{0.85\textwidth}}{\textsuperscript{3}\small Add to prompt: ``Output the single center point coordinate in [x, y] format. Scale coordinates from 0 to 1000.'' The predicted point is converted to a heatmap using Gaussian blurring for evaluation.} \\
\multicolumn{5}{p{0.85\textwidth}}{\textsuperscript{4}\small CLIPSeg predicts a segmentation mask, which is combined with the LVLM bounding box via Hadamard product to obtain the final localization heatmap.}

\end{tabular}
\end{table}

Direct LVLM prompting performs poorly for affordance grounding. 
Using the same prompt as our framework, Method A (LVLM, Qwen3-VL~\cite{bai2025qwen3}) produces only textual responses and therefore cannot generate spatial predictions. 
Method B follows the official grounding protocol of
Qwen3-VL and prompts the model to predict bounding box coordinates, achieving an NSS of 1.085. Method C instead prompts the LVLM to output the center
coordinate of the affordance region, resulting in worse
performance (NSS 0.675, KLD 9.054).
% \edit{Method D combines the predicted bounding box from the LVLM with a CLIPSeg segmentation mask via Hadamard product to obtain a pixel-level localization map. However, this refinement yields similar performance to Method B.}{} 

% \qq{Should this LVLM +CLIPseg be categorized in group2, multiple foundation models?} \lsi{Thanks. That makes sense so I 've changed them} 

%
Furthermore, combined foundation-model baselines (Methods D, E, and F) also show inferior performance compared to TokAG. 
Method D combines the predicted bounding box from the LVLM with a CLIPSeg segmentation mask via Hadamard product to obtain a pixel-level localization map. However, this refinement yields similar performance to Method B.
While vision foundation models (VFMs) like SAM 3 excel at whole-object segmentation, they struggle to localize fine-grained functional regions, which is crucial for accurate affordance grounding. 
% \edit{}{ Similarly, combining an LVLM with CLIPseg (Method D) underperforms, as text-prompted segmentation models lack the grounding capabilities required for object interactions.} 
%
Consequently, these combined foundation-model baselines are less effective at capturing precise interaction regions, yielding inferior results (e.g., NSS of 1.189 and 0.657, respectively). 
Overall, these results indicate that although current foundation models capture rich semantic knowledge and object-level visual information, naively prompting LVLMs or directly combining existing foundation models is insufficient to localize precise interaction regions for affordance grounding in a zero-shot setting.
In contrast, TokAG bypasses explicit coordinate prediction and instead extracts spatial signals directly from cross-modal attention maps, enabling accurate affordance localization. 
% \hawk{In general, the newly added discussion is good. Two minor issues: 1) pipelined baselines are less commonly used, we can just use baselines. 2) Mehtod E and F use vision foundation models, but not just vision foundation models. Foundation-model-based baselines is the term the reviewer used in their review.}

Additionally, TokAG doesn't depend on the specific segmentation model. To demonstrate this flexibility, we test our model by using SAM3 instead of CLIPSeg. While the baseline by Moon et al. achieves an NSS of 1.368, our framework with SAM3 obtains an NSS of 1.416. Although this is slightly lower than our default configuration with CLIPSeg (NSS 1.514), both variants consistently outperform the state-of-the-art. 

To further analyze the impact of segmentation quality, we evaluated TokAG on 40 complex multi-object scenes, comparing cases with successful versus failed CLIPSeg outcomes. Successful segmentations yield an NSS of 1.665, whereas failures degrade performance to an NSS of 1.168. To investigate these limitations, we also conducted an error analysis on the AGD20K Unseen (Set 2) dataset, identifying 77 failure cases out of 540 test images. Among these, CLIPSeg segmentation errors account for 37.7\% (29 cases), while token selection errors account for 62.3\% (48 cases). This distribution indicates that while a reasonable object mask enhances performance, the primary remaining challenge lies in selecting semantically relevant tokens. Nevertheless, even under segmentation failure or when utilizing alternative architectures like SAM3, TokAG maintains competitive performance, verifying that our framework is not tied to a specific segmentation model.

\begin{table}[t]
\centering
\setlength{\tabcolsep}{8pt}
\caption{
Evaluation of layer and head aggregation strategies on the AGD20K dataset (Set 2).
Methods A–B evaluate different layer or head selections,
while Method C aggregates attention across all layers and heads (ours).
}
\label{tab:model_layer_comparison}

\begin{tabular}{@{}cccccc@{}}
\toprule
& \multicolumn{1}{c}{\textbf{Type}} & \multicolumn{1}{c}{\textbf{Name}} & \textbf{KLD} $\downarrow$ & \textbf{SIM} $\uparrow$ & \textbf{NSS} $\uparrow$ \\ \midrule
(A-1) & \multirow{6}{*}{\makecell[c]{Layer \& Head \\ Selection}} & L25-H5      & 1.132          & 0.441          & 1.431          \\
(A-2) &  & L25      & 1.090          & 0.450          & 1.471          \\
(A-3) &  & H5      & 1.062         & 0.454          & 1.510          \\
(B-1) &                                                           & L16-H15      & 1.135          & 0.446          & 1.459          \\
(B-2) &                                                           & L16      & 1.100          & 0.461          & 1.466          \\
(B-3) &                                                           & H15      & 1.100          & 0.450          & 1.476          \\ \midrule

(C) &                                                           & Ours    & \textbf{1.060} & \textbf{0.455} & \textbf{1.514} \\
\bottomrule
\end{tabular}
\end{table}

\subsection{Evaluation of the Token Selection Mechanism}
% \noindent \textbf{Analysis of the spatial-aware token-selection mechanism.} 

% Our framework relies on two core operations to transform LVLM semantic signals into accurate affordance heatmaps: (i) \textbf{Layer/Head aggregation} and (ii) \textbf{spatial-aware token selection}. We evaluate multiple strategies for both operations, with quantitative results detailed in Table~\ref{tab:model_layer_comparison}.

Our framework converts the semantic knowledge of LVLMs into spatial
affordance heatmaps through two key operations:
(i) \textbf{layer/head aggregation} and
(ii) \textbf{spatial-aware token selection}.
We evaluate several alternative strategies for both components,
with results summarized in Table~\ref{tab:model_layer_comparison} and Table~\ref{tab:token_selection_comparison}.

\noindent \textbf{Layer and Head Aggregation.}
Transformer models exhibit hierarchical representations where early
layers encode spatial features while deeper layers capture semantic
information~\cite{ghiasi2022vision}. Prior works in visual grounding
often rely on selecting specific layers or attention heads
(e.g., middle layers) to extract spatial signals~\cite{kang2025your}.
However, identifying optimal layers or heads typically requires
supervision or structural pruning~\cite{michel2019sixteen}, which is
infeasible in our zero-shot setting.

We therefore compare several layer/head selection strategies
(Table~\ref{tab:model_layer_comparison}, A–B).
Empirically, restricting attention extraction to individual layers
or heads leads to inferior performance.
In contrast, aggregating attention weights across all layers and heads
(Method C) yields improved results (e.g., NSS 1.514).
This suggests that affordance localization relies on distributed
semantic–spatial cues across different layers and heads.

\begin{table}[t]
\centering
\setlength{\tabcolsep}{8pt}
\caption{
Evaluation of token selection strategies for affordance grounding on the AGD20K dataset (Set 2).
Methods D–H evaluate several heuristic token choices,
while Method I corresponds to our spatial-aware Top-1 token selection.
}
\label{tab:token_selection_comparison}
\begin{tabular}{@{}cccccc@{}}
\toprule
& \multicolumn{1}{c}{\textbf{Type}} & \multicolumn{1}{c}{\textbf{Name}} & \textbf{KLD} $\downarrow$ & \textbf{SIM} $\uparrow$ & \textbf{NSS} $\uparrow$ \\ \midrule

(D) & \multirow{6}{*}{\makecell[c]{Token \\ Selection}}         & Average All Tokens   & 1.111          & 0.435          & 1.457          \\
(E) &                                                           & First Output Token   & 1.078          & 0.444          & 1.493          \\
(F) &                                                           & Last Output Token    & 1.358          & 0.364          & 1.116          \\
(G) &                                                           & Noun Tokens Only     & 1.112          & 0.435          & 1.461          \\
(H) &                                                           & Top-5 Tokens         & 1.084          & 0.444          & 1.482          \\ \midrule
(I) &                                                           & Ours     & \textbf{1.060} & \textbf{0.455} & \textbf{1.514} \\ \bottomrule
\end{tabular}
\end{table}

\noindent \textbf{Token Selection Strategy.} 
Existing visual grounding approaches typically rely on a predefined
token, such as the \texttt{[CLS]} token in vision transformers
\cite{chefer2021transformer, wu2024faithfulness}.
However, generative LVLMs do not contain a dedicated classification
token, and therefore require alternative strategies.
Previous work has explored using the first output token
\cite{zhang2025mllms} or the last prompt token
\cite{kang2025your} for localization.
To identify the most informative token for affordance grounding,
we compare several token-selection strategies
(Table~\ref{tab:token_selection_comparison}, D–H), including:
averaging all tokens, selecting the first output token,
selecting the last output token, restricting to noun tokens,
and selecting the Top-$k$ tokens.

These strategies perform suboptimally in general.
Averaging all tokens (Method D) introduces noise from
non-informative tokens and results in spatially diffused attention maps.
Similarly, relying on fixed positional tokens
(Method E or Method F) degrades performance because these tokens
often lack explicit semantic cues related to the affordance region.

Interestingly, restricting the selection to noun tokens
(Method G) also underperforms.
We observe that affordance-related activations frequently appear in
articles (e.g., ``a'', ``the'') that precede the target noun.
For example, in the phrase ``cut with a knife,''
the token ``a'' often produces the strongest spatial activation.
This occurs because autoregressive attention for the current token
is influenced by visual features required to predict the subsequent
token, causing functional regions to be activated earlier in the
generation process.

Finally, selecting multiple tokens (Top-5, Method H)
slightly increases SIM due to broader spatial coverage,
but degrades KLD and NSS as attention spreads beyond
the primary functional region.
In contrast, our spatial-aware Top-1 token selection (Method I) instead identifies the single token whose attention map shows the strongest activation over the target object.
This mechanism effectively suppresses noise and
autoregressive token drift, achieving the best overall performance
(KLD 1.060, SIM 0.455, NSS 1.514).

\begin{figure}[t]
    \centering
    \includegraphics[width=0.98\columnwidth]{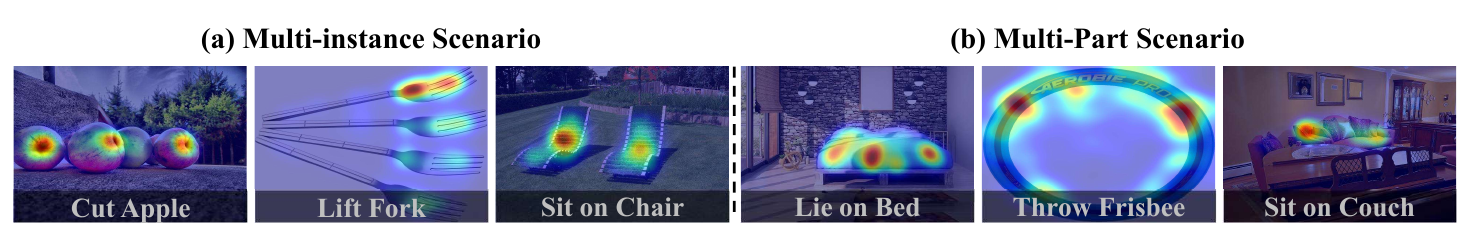}
    \caption{Examples of Multi-objects and Multi-part scenarios}
    \label{fig:multi_objects}
    \vspace{-0.3cm}
\end{figure}

\noindent \textbf{Analysis of Multi-Instance and Multi-Part Scenarios.}
Although our token-selection strategy selects a single Top-1 token, the resulting localization is not restricted to a single spatial region as in Fig.~\ref{fig:multi_objects}. The cross-modal attention map of the selected token naturally captures multiple instances or functional parts, provided they share identical affordance properties. Consequently, TokAG can localize multi-instance and multi-part scenarios.
% \hawk{The Table here is a typo. This is issue Number 1. Second problem, adding such a short subsection is not a good option in the method section.}

\vspace{-0.35cm}

% \begin{figure}[t]
%     \centering
%     \includegraphics[width=\linewidth]{figures/add_fig_cropped.pdf}
%     \caption{Representative failure cases of TokAG.
% In these examples, the predicted affordance heatmaps attend to
% incorrect or diffuse regions rather than the true functional parts.
% Such errors often occur when the affordance region is small,
% visually ambiguous, or attention mechanism is distracted by salient visual artifacts such as logos, text, or high-contrast markings on the object surface (e.g., the ``REMO'' logo on the drum).}
%     \label{fig:failurecase}
% \end{figure}

% Another common failure scenario arises when attention mechanism is distracted by salient visual artifacts such as logos, text, or high-contrast markings on the object surface (e.g., the ``REMO'' logo on the drum). This bias leads to suboptimal affordance localization, where the model attends to task-irrelevant visual patterns rather than the functional interaction regions.

% \caption{Limitations of attention-based grounding. We observe that the attention mechanism is often distracted by salient visual artifacts such as logos, text, or high-contrast markings on the object surface (e.g., the ``REMO'' logo on the drum). This bias leads to suboptimal affordance localization, where the model attends to task-irrelevant visual patterns rather than the functional interaction regions.}

\subsection{Failure cases}

% Although TokAG achieves competitive performance across multiple benchmarks, it still exhibits several failure cases. \ref{fig:failurecase} shows representative examples where the predicted affordance heatmaps deviate from the true functional regions. These failures typically occur when the target affordance region is small, visually ambiguous, or partially occluded, making it difficult for the model to isolate the precise interaction location.
% Another common failure scenario arises when the attention mechanism is distracted by salient visual artifacts such as logos, text, or high-contrast markings on the object surface (e.g., the ``REMO'' logo on the drum). This bias leads to suboptimal affordance localization, where the model attends to task-irrelevant visual patterns rather than the functional interaction regions. Such behavior suggests that the attention responses extracted from LVLMs may sometimes prioritize visually distinctive cues over functional affordance signals, especially when the interaction region itself lacks strong visual contrast.

\begin{figure}[t]
    \centering
    \includegraphics[width=\linewidth]{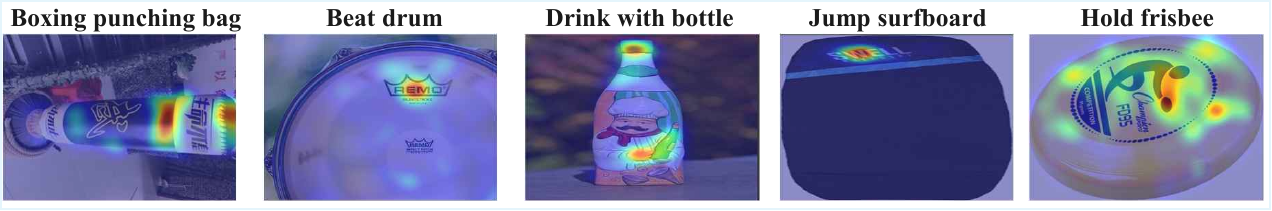}
    \caption{Representative failure cases of TokAG.
In these examples, the predicted affordance heatmaps attend to
incorrect or diffuse regions rather than the true functional parts.
Such errors often occur when the affordance region is small,
visually ambiguous, or attention mechanism is distracted by salient visual artifacts such as logos, text, or high-contrast markings on the object surface (e.g., the ``REMO'' logo on the drum).}
    \label{fig:failurecase}
\end{figure}

Fig.~\ref{fig:failurecase} shows representative failure cases where the predicted affordance heatmaps deviate from the true functional regions. These failures typically occur when the target affordance region is small, visually ambiguous, or partially occluded, or when salient visual artifacts (e.g., logos or text) distract the attention mechanism from the true functional regions.

\subsection{Limitations and Future Directions}

% While TokAG achieves promising performance across multiple benchmarks,
% it also has several limitations that suggest directions for future work.
% First, our framework operates in a zero-shot and training-free
% setting. Although this design enables generalization without
% requiring supervision, it also prevents the model from leveraging
% dataset-specific signals, such as exocentric training images or weak supervision available in affordance datasets. Future work could
% explore integrating lightweight adaptation modules or self-supervised
% learning strategies to further improve affordance localization.

% Finally, TokAG assumes that the target affordance can be represented by a single semantic token. This assumption is effective for most affordance categories in existing benchmarks, but is insufficient for 

Although TokAG performs well on existing benchmarks, it currently assumes that the target affordance can be represented by a single semantic token. This assumption aligns well with the object-centric affordance definitions used in current benchmarks, but may be insufficient for compositional affordances (e.g., ``pour from'' requiring both the handle and the spout) and bimanual interactions (e.g., opening a jar with one hand holding the jar and the other twisting the lid) that require simultaneous grounding of multiple functional regions. 
% The performance degradation observed with multi-token aggregation (Table~\ref{tab:token_selection_comparison}) further suggests that simply combining multiple tokens is insufficient for modeling such interactions.
%
Developing multi-region reasoning and token aggregation strategies for compositional affordances is an important direction for future work.

\section{Conclusion}
\label{sec:conclusion}

% We introduced \textbf{TokAG}, a novel zero-shot framework for affordance grounding, boosting action-specific localization without the need for external supervision. \textbf{TokAG} exploits token-level semantic-spatial signals to transform the implicit action semantics of Large Vision-Language Models (LVLMs) into precise pixel-level heatmaps. 
% %
% In this work, we have demonstrated that the implicit semantic signals of LVLMs can be explicitly transformed into an affordance heatmaps. Rather than relying on noisy or arbitrary attention maps, our spatial-aware token-selection mechanism systematically isolates the output tokens that exhibit dominant activation over the target area.
% %
% \textbf{TokAG} improves the NSS metric by 16.8\% on the unseen split of AGD20K and by 29.7\% on HICO-IIF, consistently outperforming prior weakly supervised approaches and establishing a new state-of-the-art in zero-shot affordance grounding.

We introduced \textbf{TokAG}, a novel zero-shot framework for affordance grounding that enables action-specific localization without requiring external supervision. TokAG leverages token-level semantic–spatial signals within large vision-language models (LVLMs) to transform their implicit action semantics into pixel-level affordance heatmaps.
Our analysis shows that the spatial activations of LVLM output tokens contain meaningful cues for affordance localization. Rather than relying on noisy or arbitrary attention maps, TokAG employs a spatial-aware token selection mechanism that systematically identifies the output token whose attention map exhibits the strongest activation over the target object.
Extensive experiments demonstrate that TokAG significantly improves affordance grounding performance. In particular, it improves the NSS metric by \textbf{10.7\%} on the unseen split of AGD20K and by \textbf{29.7\%} on HICO-IIF, consistently outperforming prior weakly supervised approaches and establishing a new state-of-the-art for zero-shot affordance grounding.

\noindent \textbf{Acknowledgements.}
% We would like to express our sincere gratitude to the Area Chair and the anonymous reviewers for their constructive comments and valuable suggestions, which greatly helped improve the quality of this paper.
We thank the anonymous reviewers and area chairs for their constructive feedback, which helped improve this paper.

% \makeatletter
% \let\@oldmaketitle\@maketitle% Store \@maketitle
% \renewcommand{\@maketitle}{\@oldmaketitle% Update \@maketitle to insert...
% \input{figures/qualitative_results_ablation_v1}}
% \makeatother

\bibliographystyle{splncs04}
\bibliography{main}

% \nopagebreak

\input{supp.tex}

\end{document}

%% file: figures/qualitative_results_v2.tex
%\begin{figure*}
\begin{center}
    \centering

    % Add action label
    \begin{minipage}[c]{0.18\linewidth}
         %\vspace{0.1em}
		\centerline{\scriptsize{Action: sit on}}
	\end{minipage}
    \begin{minipage}[c]{0.18\linewidth}
        %\vspace{0.1em}
		\centerline{}
		%\vspace{0.3em}
	\end{minipage}
	\begin{minipage}[c]{0.18\linewidth}
        %\vspace{0.1em}
		\centerline{}
		%\vspace{0.3em}
	\end{minipage}	
    \begin{minipage}[c]{0.18\linewidth}
        %\vspace{0.1em}
		\centerline{}
		%\vspace{0.3em}
	\end{minipage}
    \begin{minipage}[c]{0.18\linewidth}
        %\vspace{0.1em}
		\centerline{}
		%\vspace{0.3em}
	\end{minipage}

    % First row
    \begin{minipage}[c]{0.18\linewidth}
		\includegraphics[width=\linewidth]{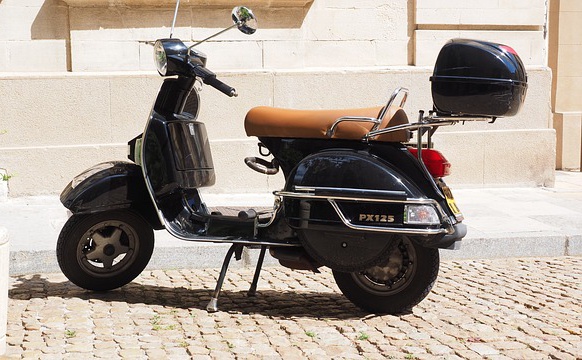}
	\end{minipage}
    \begin{minipage}[c]{0.18\linewidth}
		\includegraphics[width=\linewidth]{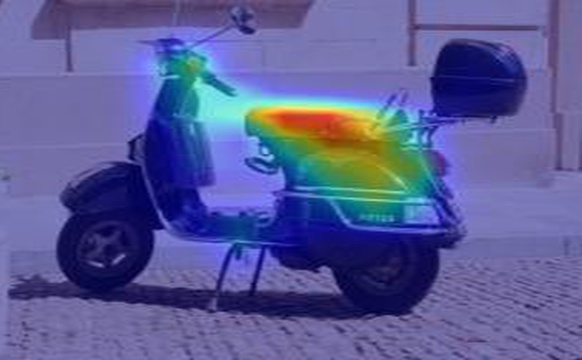}
	\end{minipage}
	\begin{minipage}[c]{0.18\linewidth}
		\includegraphics[width=\linewidth]{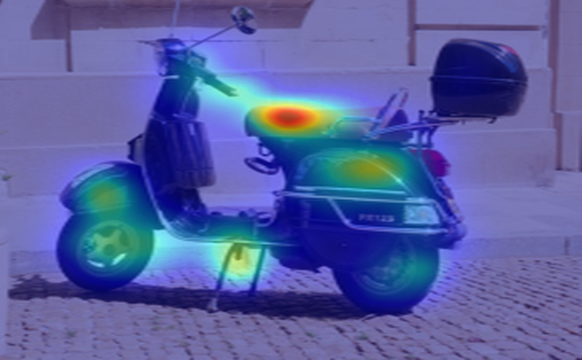}
	\end{minipage}
	\begin{minipage}[c]{0.18\linewidth}
		\includegraphics[width=\linewidth]{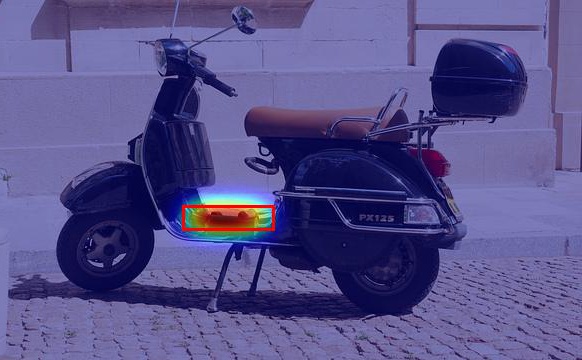}
	\end{minipage}    
    \begin{minipage}[c]{0.18\linewidth}
		\includegraphics[width=\linewidth]{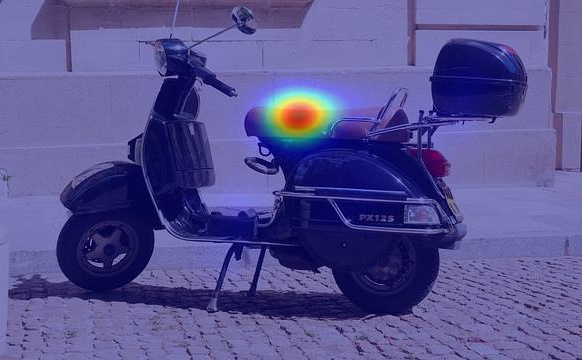}
	\end{minipage}

    % Add action label
    \begin{minipage}[c]{0.18\linewidth}
         %\vspace{0.1em}
		\centerline{\scriptsize{Action: push}}
	\end{minipage}
    \begin{minipage}[c]{0.18\linewidth}
        %\vspace{0.1em}
		\centerline{}
		%\vspace{0.3em}
	\end{minipage}
	\begin{minipage}[c]{0.18\linewidth}
        %\vspace{0.1em}
		\centerline{}
		%\vspace{0.3em}
	\end{minipage}	
    \begin{minipage}[c]{0.18\linewidth}
        %\vspace{0.1em}
		\centerline{}
		%\vspace{0.3em}
	\end{minipage}
    \begin{minipage}[c]{0.18\linewidth}
        %\vspace{0.1em}
		\centerline{}
		%\vspace{0.3em}
	\end{minipage}

    % Second row 
    \begin{minipage}[c]{0.18\linewidth}
		\includegraphics[width=\linewidth]{figures/qualitative_results/motorcycle/motorcycle_000592.jpg}
	\end{minipage}
    \begin{minipage}[c]{0.18\linewidth}
		\includegraphics[width=\linewidth]{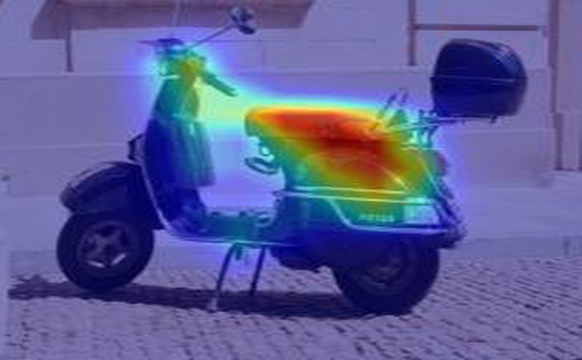}
	\end{minipage}
	\begin{minipage}[c]{0.18\linewidth}
		\includegraphics[width=\linewidth]{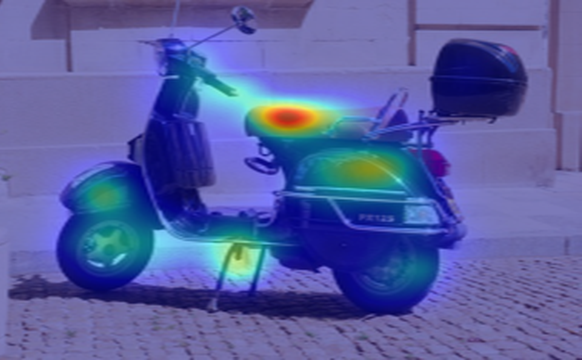}
	\end{minipage}
	\begin{minipage}[c]{0.18\linewidth}
		\includegraphics[width=\linewidth]{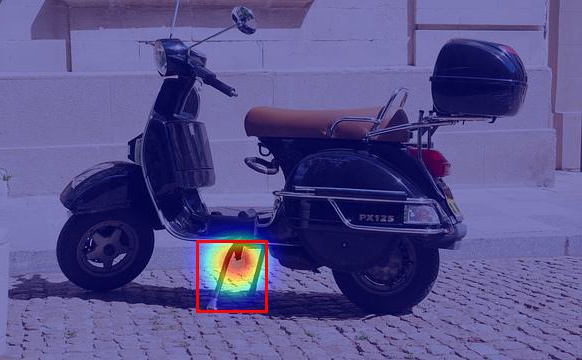}
	\end{minipage}
    \begin{minipage}[c]{0.18\linewidth}
		\includegraphics[width=\linewidth]{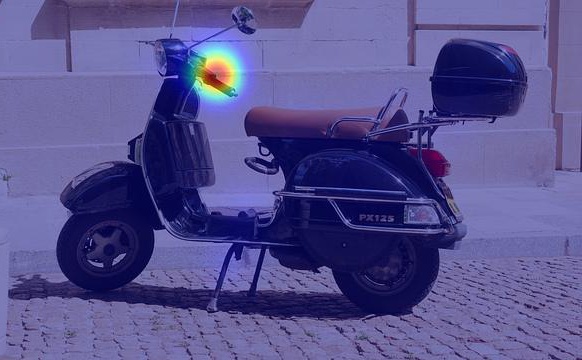}
	\end{minipage}
    
   % Second row    
    % Add action label
    \begin{minipage}[c]{0.18\linewidth}
         %\vspace{0.1em}
		\centerline{\scriptsize{Action: hold}}
	\end{minipage}
    \begin{minipage}[c]{0.18\linewidth}
        %\vspace{0.1em}
		\centerline{}
		%\vspace{0.3em}
	\end{minipage}
	\begin{minipage}[c]{0.18\linewidth}
        %\vspace{0.1em}
		\centerline{}
		%\vspace{0.3em}
	\end{minipage}	
    \begin{minipage}[c]{0.18\linewidth}
        %\vspace{0.1em}
		\centerline{}
		%\vspace{0.3em}
	\end{minipage}
    \begin{minipage}[c]{0.18\linewidth}
        %\vspace{0.1em}
		\centerline{}
		%\vspace{0.3em}
	\end{minipage}
    
    % third row 
    \begin{minipage}[c]{0.18\linewidth}
		\includegraphics[width=\linewidth]{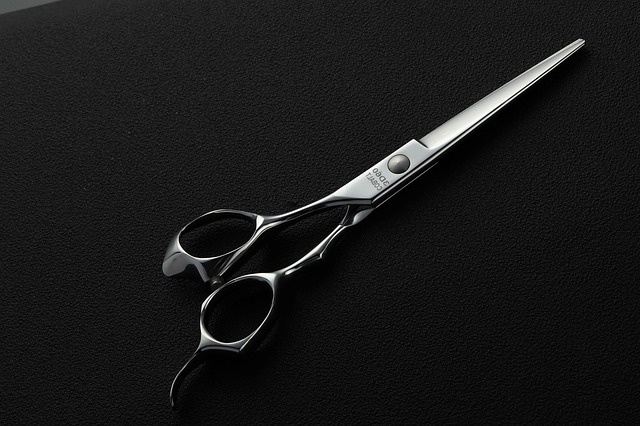}
	\end{minipage}
    \begin{minipage}[c]{0.18\linewidth}
		\includegraphics[width=\linewidth]{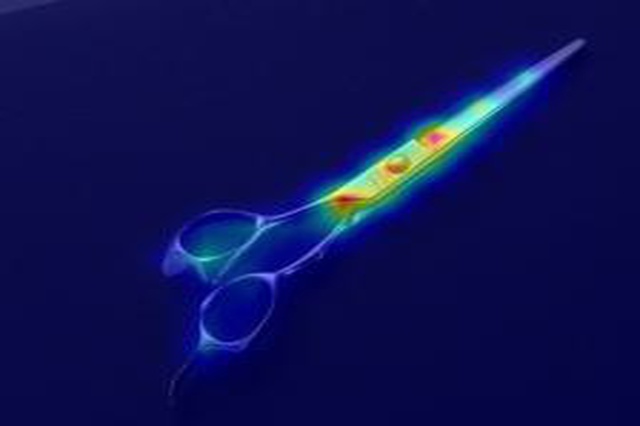}
	\end{minipage}
	\begin{minipage}[c]{0.18\linewidth}
		\includegraphics[width=\linewidth]{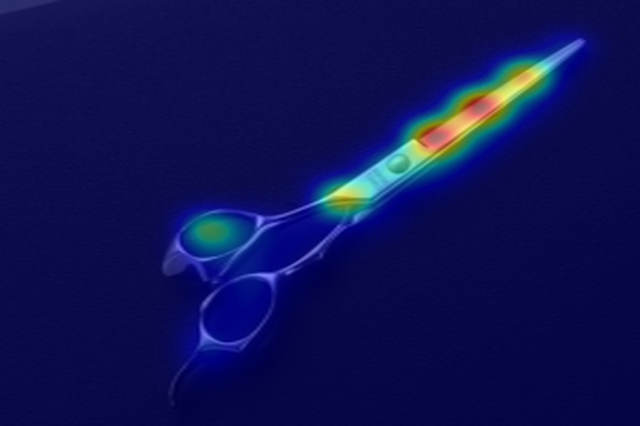}
	\end{minipage}
    \begin{minipage}[c]{0.18\linewidth}
		\includegraphics[width=\linewidth]{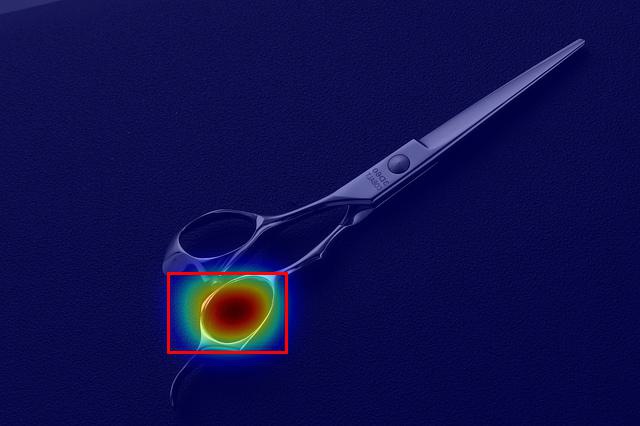}
	\end{minipage}
    \begin{minipage}[c]{0.18\linewidth}
		\includegraphics[width=\linewidth]{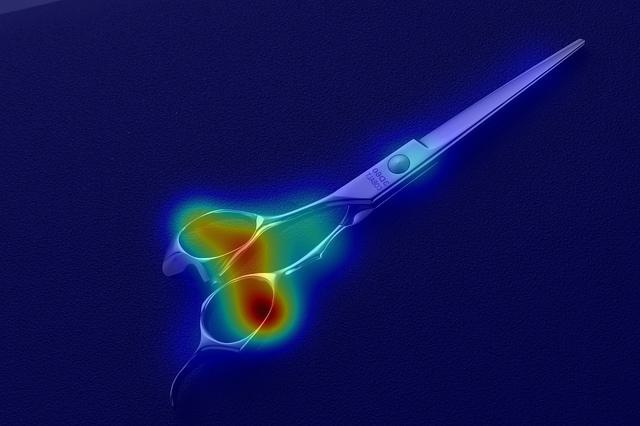}
	\end{minipage}

    % Add action label
    \begin{minipage}[c]{0.18\linewidth}
         %\vspace{0.1em}
		\centerline{\scriptsize{Action: cut with}}
	\end{minipage}
    \begin{minipage}[c]{0.18\linewidth}
        %\vspace{0.1em}
		\centerline{}
		%\vspace{0.3em}
	\end{minipage}
	\begin{minipage}[c]{0.18\linewidth}
        %\vspace{0.1em}
		\centerline{}
		%\vspace{0.3em}
	\end{minipage}	
    \begin{minipage}[c]{0.18\linewidth}
        %\vspace{0.1em}
		\centerline{}
		%\vspace{0.3em}
	\end{minipage}
    \begin{minipage}[c]{0.18\linewidth}
        %\vspace{0.1em}
		\centerline{}
		%\vspace{0.3em}
	\end{minipage}
    
    % second row 
     \begin{minipage}[c]{0.18\linewidth}
		\includegraphics[width=\linewidth]{figures/qualitative_results/sissors/scissors_003229.jpg}
    \end{minipage}
     \begin{minipage}[c]{0.18\linewidth}
		\includegraphics[width=\linewidth]{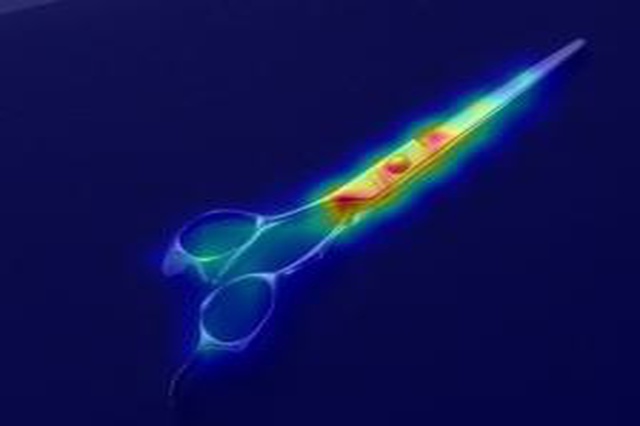}
	\end{minipage}
    \begin{minipage}[c]{0.18\linewidth}
		\includegraphics[width=\linewidth]{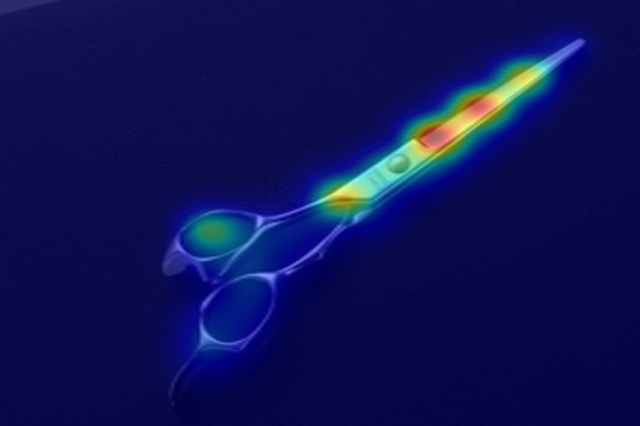}
	\end{minipage}
    \begin{minipage}[c]{0.18\linewidth}
		\includegraphics[width=\linewidth]{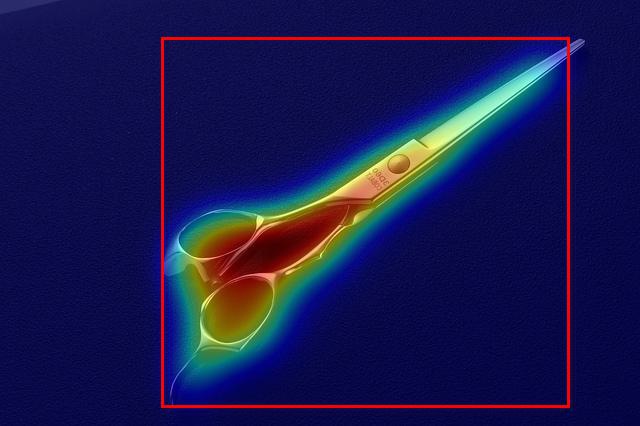}
	\end{minipage}
    \begin{minipage}[c]{0.18\linewidth}
		\includegraphics[width=\linewidth]{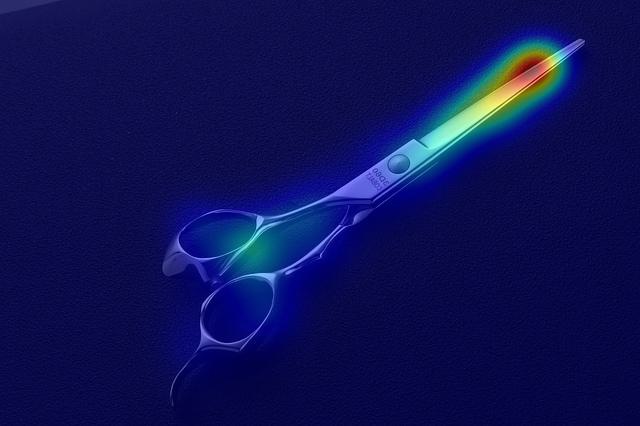}
	\end{minipage}

     % Paper label
    \begin{minipage}[c]{0.18\linewidth}
         %\vspace{0.1em}
		\centerline{\scriptsize{Input Image}}
	\end{minipage}
    \begin{minipage}[c]{0.18\linewidth}
        %\vspace{0.1em}
		\centerline{\scriptsize{LOCATE}}
		%\vspace{0.3em}
	\end{minipage}
	\begin{minipage}[c]{0.18\linewidth}
        %\vspace{0.1em}
		\centerline{\scriptsize{WSMA}}
		%\vspace{0.3em}
	\end{minipage}	
    \begin{minipage}[c]{0.18\linewidth}
        %\vspace{0.1em}
		\centerline{\scriptsize{LVLM}}
		%\vspace{0.3em}
	\end{minipage}
    \begin{minipage}[c]{0.18\linewidth}
    %\vspace{0.1em}
    \centerline{\scriptsize{Ours}}
    %\vspace{0.3em}
    \end{minipage}

\vspace{-0.2em}
\captionof{figure}{
Comparison of affordance grounding results.
LOCATE~\cite{li2023locate} struggles with action co-occurrence in exocentric images,
while WSMA~\cite{xu2024weakly} fails to distinguish semantically similar actions due to its reliance on short action phrases.
Directly prompting a large vision–language model (LVLM, Qwen3-VL~\cite{bai2025qwen3}) also produces inaccurate localization.
% Since LVLMs do not directly output pixel-wise heatmaps, the results are obtained using CLIPSeg~\cite{luddecke2022image} masks. 
Since LVLMs do not output heatmaps, the results are obtained using CLIPSeg~\cite{luddecke2022image}. 
In contrast, our method produces more accurate affordance localization.
% Prior works often fail to provide precise affordance maps: LOCATE~\cite{li2023locate} struggles with action co-occurrence in exocentric images, while WSMA~\cite{xu2024weakly} fails to distinguish semantically similar actions due to its reliance on brief action phrases. While Large Vision-Language Models are advanced in text generation, they have limitations in localizing their results to semantic action regions. \edit{}{Comment: To obtain results from existing LVLMs, we extracted bounding box coordinates from the LVLMs and applied object-name-based CLIPSeg masks to generate the results in the fourth column.]} In contrast, our method captures rich semantic details from LVLMs, transform its implicit semantic signals into affordance heatmaps. 
}
\label{fig:representative_result}
% \vspace{-1mm}
\end{center}

%% file: supp.tex
\makeatletter
\textsubscript{}
\setcounter{section}{0}
\renewcommand{\thesection}{\Alph{section}}
\addtocounter{section}{0} 
\makeatother

\setcounter{figure}{0}
\setcounter{table}{0}
\renewcommand{\thefigure}{S\arabic{figure}}
\renewcommand{\thetable}{S\arabic{table}}

\clearpage
\nopagebreak
\begin{center}
    \makeatletter
    {\LARGE \bf Supplementary Material for \\ \vspace{0.2em}
    Token-Based Affordance Grounding with \\ \vspace{0.2em}
    Large Vision-Language Models \par}
    \makeatother
\end{center}
\vspace{2.5em}

\vspace{-0.5em}

\section{Implementation Details}
\label{sec:implementation}

In this section, we provide additional implementation details of the proposed tokAG, our zero-shot affordance grounding framework. Specifically, we elaborate on the LVLM inference settings, the object-masking pipeline, and the mathematical formulations for our token-centric attention extraction and post-processing techniques.

\subsection{LVLM Configuration and Inference Settings}
We employ Qwen3-VL~\cite{bai2025qwen3} (2B and 32B) as our LVLM (Large Vision Language Model) backbones. To facilitate the extraction of internal cross-attention maps, the models are instantiated with an eager attention implementation (i.e., \texttt{attn\_implementation="eager"}), and we perform inference using \texttt{bfloat16} precision. The input images are processed by Qwen's vision encoder, which maps the image into a spatial vision grid.
To utilize implicit semantic signals in LVLMs, we prompt the model with the following template:
\begin{quote}
\textit{``When people perform \{action\} with \{object\_name\}, which part of the \{object\_name\} is used for `\{action\}'? Answer in one sentence.''}
\end{quote}

To constrain the predicted affordance regions within the target object, we employ CLIPSeg~\cite{luddecke2022image} as an off-the-shelf segmenter. By providing the target object's name as a text prompt, the generated raw output logits are passed through a sigmoid activation function to obtain continuous probability maps. These maps are resized using interpolation and thresholded to form the binary object mask $M_{obj}$.

As formulated in Sec.~3.1 (main paper), during the LVLM generation, we extract the cross-attention maps corresponding to the generated text tokens. Based on the CLIPSeg-generated mask $M_{obj}$, we select the Top-1 token $t^*$ that yields the highest spatial activation within the target object area ($M_{obj}$). The aggregated attention map of this selected token serves as the raw affordance heatmap. Initially at the LVLM's vision grid resolution ($H_{grid} \times W_{grid}$), this heatmap is finally bilinearly upsampled to match the fixed input image size of $1000 \times 1000$.

\subsection{Attention Map Post-Processing}

Although the selected attention map $\mathcal{H}$ is constrained within the target object boundaries, the raw activations can sometimes be overly concentrated or exhibit peaks. This concentration is often further amplified by the ``attention sink'' phenomenon~\cite{kang2025see}, where a few specific visual patches disproportionately absorb attention weights, thereby suppressing other semantically relevant regions. To mitigate this effect and ensure that the high-response regions adequately cover the entire functional parts, we apply a spatial refinement step as follows.

First, the upsampled and masked heatmap $\mathcal{H}$ is normalized by its maximum value. We then apply a fractional power scaling of $0.75$ to the heatmap:

$$ \mathcal{H}_{scaled} = \mathcal{H}^{0.75} $$

This fractional scaling ($p < 1$) effectively softens overly sparse and highly concentrated attention patches, amplifying weak but relevant interaction cues to form a more cohesive functional region.

Finally, to ensure spatial continuity and soften rigid boundaries, we apply an adaptive Gaussian blur. Since the final heatmap must be overlaid onto the original image for affordance visualization, we maintain scale invariance by dynamically calculating the standard deviation $\sigma$ and the kernel size $K$ based on the original image dimensions ($W_{orig} \times H_{orig}$), rather than the fixed $1000 \times 1000$ resolution used during LVLM inference:
$$ \sigma = \min(W_{orig}, H_{orig}) \times 0.05 $$
$$ K = 2 \times \lfloor 3\sigma \rfloor + 1 $$
Following the blurring operation, the heatmap undergoes a final min-max normalization to yield the ultimate affordance probability distribution, ranging strictly between 0 and 1.

\section{Additional Ablation Studies}
\label{sec:ablation}

\subsection{Effect of Input Prompt Design}
\label{subsec:effect_of_prompt}

In our framework, the LVLM's internal cross-attention map is conditioned on the generated textual response, which is driven by the input prompt. To evaluate the robustness of our token-centric attention extraction mechanism under different instructions, we conducted an ablation study using 11 different prompts. As detailed in Table \ref{tab:prompt_design}, we varied both the system instructions (ranging from a standard ``helpful assistant'' to an ``analytical computer vision model'') and the specific phrasing of the user query. 

The quantitative results demonstrate the stability of our approach. Despite variations in the linguistic structure and tone of the input prompts, the affordance grounding performance remains comparable across different instructions. Assigning an analytical role to the LVLM, such as ``an AI vision analyzer'' (Case 1) or ``a task-oriented visual AI'' (Case 2), leads to slight performance improvements (e.g., KLD of 1.039 for Case 1). However, the ``Original'' prompt, which specifies the action-object relationship without strict formatting constraints, also achieves competitive results. We therefore use it as the default configuration in the main experiments to demonstrate that the proposed framework works well without relying on extensive prompt engineering.

\small
\renewcommand{\arraystretch}{1.4}
\begin{xltabular}{\textwidth}{@{} c p{2.5cm} X ccc @{}}
\caption{Ablation study on the effect of different input prompt designs on the AGD20K dataset (Set 2). We vary the system instructions and user queries to evaluate the robustness of our affordance grounding framework. Cases 1--3 include a single-sentence length constraint, whereas Cases 4--10 remove this constraint. The best performance is \textbf{bolded} and the second-best is \underline{underlined}.}
\label{tab:prompt_design} \\ 

\toprule
\textbf{ID} & \textbf{System Prompt} & \textbf{User Prompt} & \textbf{KLD} $\downarrow$ & \textbf{SIM} $\uparrow$ & \textbf{NSS} $\uparrow$ \\
\midrule
\endfirsthead

\multicolumn{6}{c}{{\tablename\ \thetable{} -- continued from previous page}} \\
\toprule
\textbf{ID} & \textbf{System Prompt} & \textbf{User Prompt} & \textbf{KLD} $\downarrow$ & \textbf{SIM} $\uparrow$ & \textbf{NSS} $\uparrow$ \\
\midrule
\endhead

\midrule
\multicolumn{6}{r}{{Continued on next page...}} \\
\endfoot

\bottomrule
\endlastfoot

Original & You are a helpful language and vision assistant. & When people perform \{action\} with \{object\_name\}, which part of the \{object\_name\} is used for `\{action\}'? answer in one sentence. & 1.060 & 0.455 & 1.514 \\

Case\_1 & You are an AI vision analyzer. & Identify the exact part of the \{object\_name\} that people use to `\{action\}'. Keep your response to one sentence. & \textbf{1.039} & \underline{0.461} & \textbf{1.532} \\

Case\_2 & You are a task-oriented visual AI. & Your task is to identify the part of the \{object\_name\} used for `\{action\}'. Provide the answer in one sentence. & \underline{1.057} & \textbf{0.461} & 1.509 \\

Case\_3 & You are a simple and clear AI assistant. & To `\{action\}', what part of the \{object\_name\} do you need? Give me a one-sentence answer. & 1.067 & 0.451 & \underline{1.522} \\

Case\_4 & You are an AI vision analyzer. & Identify the exact part of the \{object\_name\} that people use to `\{action\}'. & 1.107 & 0.452 & 1.453 \\

Case\_5 & You are a task-oriented visual AI. & Your task is to identify the part of the \{object\_name\} used for `\{action\}'. & 1.102 & 0.455 & 1.472 \\

Case\_6 & You are a simple and clear AI assistant. & To `\{action\}', what part of the \{object\_name\} do you need? & 1.125 & 0.439 & 1.437 \\

Case\_7 & You are a spatial-aware vision assistant. & Locate and describe the specific physical area on the \{object\_name\} where the `\{action\}' takes place. & 1.082 & 0.456 & 1.472 \\

Case\_8 & You are an action-recognition specialist. & Focusing on the action `\{action\}', detail the exact component of the \{object\_name\} that affords this interaction. & 1.162 & 0.437 & 1.383 \\

Case\_9 & You are a direct and helpful assistant. & What is the primary part of the \{object\_name\} that is required to `\{action\}'? & 1.087 & 0.449 & 1.481 \\

Case\_10 & You are an analytical AI that understands object affordances. & Analyze the relationship between the action `\{action\}' and the \{object\_name\}. Which part of the object is manipulated or interacted with? & 1.129 & 0.438 & 1.440 \\

\end{xltabular}

In addition, we investigate the impact of output length constraints on attention quality. When comparing Cases 1--3 against their unconstrained counterparts (Cases 4--6), a performance difference is observed. For instance, Case 1 incorporates the directive, ``Keep your response to one sentence,'' whereas Case 4 uses the same instruction (``Identify the exact part of the \{object\_name\} that people use to `\{action\}'.'') but omits the length constraint. As shown in Table~\ref{tab:prompt_design}, the unconstrained prompts show lower performance across the metrics (e.g., a KLD of 1.107 for Case 4 versus 1.039 for Case 1). We attribute this to the tendency of LVLMs to generate longer, descriptive responses when unconstrained at inference time. This added length introduces linguistic noise, diluting the cross-modal attention weights across task-irrelevant words and background visual patches. By applying a single-sentence limit, the model is guided to produce a concise and semantically focused response. Consequently this length constraint helps suppress background noise and allows the model's internal cross-attention to focus on the target affordance region, yielding improved spatial grounding.

Beyond optimizing individual prompts through such constraints, we also note the possibility of a multi-prompt ensemble strategy~\cite{jiang2023calibrating}. By aggregating the cross-attention maps generated from multiple distinct prompts for a single image--action pair, it may be possible to further stabilize the predictions and achieve additional performance gains. However, while effective as an engineering technique, this ensemble approach requires executing the LVLM inference multiple times per sample, which linearly increases the computational cost and inference latency. Because the core focus of our work lies in the zero-shot token-centric attention extraction framework itself, and a single generic prompt already achieves competitive performance efficiently, we do not include the multi-prompt ensemble. We discuss it here as a possible extension for scenarios where accuracy is prioritized over computational efficiency.

\subsection{Effect of Layer and Head Selection}
\label{subsec:layer_head_selection}

In the main paper, Table 5 presents an ablation study comparing our proposed aggregation method against specific layer and head selection strategies (Methods A and B). The rationale for isolating particular combinations, such as Layer 25 Head 5 (L25-H5) and Layer 16 Head 15 (L16-H15), is inspired by recent findings from Kang \etal~\cite{kang2025your}, which suggest that large vision-language models only require a select few attention heads for image segmentation tasks.

To systematically identify these optimal heads within our framework, we extracted the internal cross-attention maps from all individual layers and heads during the LVLM generation process. We then computed a spatial similarity score by comparing each head's attention map against the explicit object mask generated by CLIPSeg. The specific layer-head combinations (e.g., L25-H5 and L16-H15) were chosen because they yielded the maximum similarity scores, indicating that these selected heads focus on the target object within the image.

However, as discussed in the main paper, relying on a single optimal layer or head is often insufficient for precise affordance grounding. Through our analysis of the cross-modal attention maps between generated output tokens and visual patches, two key observations emerge: first, spatial activations across different layers and heads are very diverse, often focusing on different image regions; second, individual attention responses can be diffuse. Despite this variance, we observe that the activations falling within the target object boundary tend to correlate with the specific functional parts associated with the queried action (e.g., highlighting the mattress for the action ``sit on bed''). Because these attention maps originate from action-conditioned output tokens generated by the LVLM, they contain rich and diverse functional cues. Therefore, our proposed approach (Method C, Table 5 main paper), which aggregates attention across all layers and heads, effectively synthesizes this diverse distribution. As demonstrated in Table 5 in the main paper, this comprehensive aggregation helps mitigate diffuse background noise and captures the nuanced semantic consensus required to pinpoint zero-shot affordance regions, outperforming single-head or single-layer selections.

To further substantiate the necessity of our aggregation approach, we conducted an experiment evaluating all possible partial aggregation strategies. Specifically, as detailed in Table~\ref{tab:all_partial_aggregation}, we investigated the performance of layer-wise aggregation (averaging attention maps across all heads within each individual layer) and head-wise aggregation (averaging across all layers for each individual head). While certain partial aggregations may occasionally show competitive metrics, they generally fall short of the robust spatial grounding achieved by comprehensively aggregating across both dimensions. We acknowledge that exhaustively searching for and optimizing a highly specific subset of layers and heads could theoretically yield marginal performance gains. However, the optimal configuration is sensitive to the specific LVLM architecture and the given image distribution. More importantly, determining such an optimal subset requires task-specific ground-truth annotations or validation sets to guide the selection process. Relying on such data-driven heuristic engineering contradicts the premise of our zero-shot setting. Therefore, our aggregation across all layers and heads serves as a robust, model-agnostic strategy, eliminating the need for manual tuning while maintaining the zero-shot nature of our framework.

\small
\renewcommand{\arraystretch}{1.2}
\begin{longtable}{@{} l l c c c @{}}
\caption{Comprehensive performance comparison of various partial aggregation strategies on the AGD20K dataset (Set 2). The table reports the results of aggregating all heads within each layer (Layer-wise) and aggregating all layers for each head (Head-wise). Within each category (Layer-wise, Head-wise, and Ours), the best performance is \textbf{bolded} and the second-best is \underline{underlined}.}
\label{tab:all_partial_aggregation} \\

\toprule
\textbf{Aggregation Strategy} & \textbf{Configuration} & \textbf{KLD} $\downarrow$ & \textbf{SIM} $\uparrow$ & \textbf{NSS} $\uparrow$ \\
\midrule
\endfirsthead

\multicolumn{5}{c}{{\tablename\ \thetable{} -- continued from previous page}} \\
\toprule
\textbf{Aggregation Strategy} & \textbf{Configuration} & \textbf{KLD} $\downarrow$ & \textbf{SIM} $\uparrow$ & \textbf{NSS} $\uparrow$ \\
\midrule
\endhead

\midrule
\multicolumn{5}{r}{{Continued on next page...}} \\
\endfoot

\bottomrule
\endlastfoot

% ==========================================
% 1. Layer-wise Aggregation Section
% ==========================================
\multicolumn{5}{@{}l}{\textbf{Layer-wise (Aggregating All Heads)}} \\
\midrule
& Layer 1  & 1.303 & 0.388 & 1.175 \\
& Layer 2  & 1.389 & 0.369 & 1.060 \\
& Layer 3  & 1.288 & 0.404 & 1.207 \\
& Layer 4  & 1.266 & 0.397 & 1.204 \\
& Layer 5  & 1.245 & 0.409 & 1.251 \\
& Layer 6  & 1.335 & 0.390 & 1.124 \\
& Layer 7  & 1.172 & 0.442 & 1.354 \\
& Layer 8  & 1.248 & 0.406 & 1.227 \\
& Layer 9  & 1.218 & 0.412 & 1.278 \\
& Layer 10 & 1.163 & 0.428 & 1.349 \\
& Layer 11 & 1.300 & 0.401 & 1.150 \\
& Layer 12 & 1.080 & 0.424 & 1.489 \\
& Layer 13 & 1.121 & 0.439 & 1.413 \\
& Layer 14 & 1.105 & 0.459 & 1.459 \\
& Layer 15 & 1.100 & 0.456 & 1.459 \\
& Layer 16 & \textbf{1.060} & 0.454 & 1.508 \\
& Layer 17 & 1.100 & \textbf{0.461} & 1.466 \\
& Layer 18 & 1.109 & 0.456 & 1.446 \\
& Layer 19 & 1.087 & 0.447 & 1.466 \\
& Layer 20 & \underline{1.065} & 0.453 & \textbf{1.514} \\
& Layer 21 & 1.070 & 0.449 & 1.502 \\
& Layer 22 & 1.067 & 0.445 & \underline{1.511} \\
& Layer 23 & 1.066 & 0.451 & 1.495 \\
& Layer 24 & 1.070 & \underline{0.460} & 1.492 \\
& Layer 25 & 1.066 & 0.447 & 1.498 \\
& Layer 26 & 1.089 & 0.450 & 1.471 \\
& Layer 27 & 1.093 & 0.447 & 1.464 \\
& Layer 28 & 1.318 & 0.390 & 1.152 \\
\midrule
% ==========================================
% 2. Head-wise Aggregation Section
% ==========================================
\multicolumn{5}{@{}l}{\textbf{Head-wise (Aggregating All Layers)}} \\
\midrule
& Head 1   & 1.142 & 0.429 & 1.388 \\
& Head 2   & 1.099 & 0.445 & 1.445 \\
& Head 3   & 1.086 & 0.455 & 1.483 \\
& Head 4   & 1.104 & 0.442 & 1.462 \\
& Head 5   & \underline{1.061} & 0.454 & \underline{1.507} \\
& Head 6   & 1.063 & 0.454 & \textbf{1.513} \\
& Head 7   & 1.077 & 0.455 & 1.469 \\
& Head 8   & 1.105 & 0.438 & 1.437 \\
& Head 9   & 1.078 & 0.442 & 1.500 \\
& Head 10  & 1.101 & 0.443 & 1.447 \\
& Head 11  & 1.143 & 0.430 & 1.403 \\
& Head 12  & 1.070 & 0.459 & 1.506 \\
& Head 13  & 1.063 & 0.458 & 1.499 \\
& Head 14  & 1.079 & \underline{0.461} & 1.505 \\
& Head 15  & \textbf{1.061} & \textbf{0.465} & 1.501 \\
& Head 16  & 1.100 & 0.450 & 1.476 \\

\midrule
% ==========================================
% 3. Ours Section
% ==========================================
\textbf{Ours -- Method C} & \textbf{All Layers \& Heads} & \textbf{1.060} & \textbf{0.455} & \textbf{1.514} \\

\end{longtable}

\section{Additional Quantitative Results}
\label{sec:quantitative_results}

\subsection{Generalization to Different LVLM Backbones}

In the main paper, we demonstrated the zero-shot affordance grounding capabilities of our framework primarily utilizing the Qwen3-VL~\cite{bai2025qwen3} backbone. To investigate whether our token-centric attention extraction mechanism generalizes across different Large Vision-Language Models (LVLMs), we conducted an additional experiment where we replace the backbone with InternVL3~\cite{zhu2025internvl3} (2B parameters). 

As shown in Table~\ref{tab:effect_of_lvlm}, while Qwen3-VL (2B) achieves slightly better performance, our framework with InternVL3 (2B) still surpasses previous state-of-the-art weakly supervised methods, demonstrating good generalization across different LVLM backbones. Specifically, on AGD20K-Set 2 (Unseen), our method with InternVL3 (2B) achieves a KLD of 1.183 and an SIM of 0.429, outperforming recent baselines such as LoopTrans (KLD 1.247) and Moon et al. (KLD 1.243). This indicates that our zero-shot approach is not strictly tied to a specific architecture and can leverage the implicit semantic signals of different LVLMs.

\begin{table}[h]
\caption{Performance comparison on the AGD20K-Set 2 (Unseen) dataset using different LVLM backbones. The best performance is \textbf{bolded} and the second-best is \underline{underlined}.}
\label{tab:effect_of_lvlm}
\centering
\renewcommand{\arraystretch}{1.2}
\begin{tabular}{@{} l >{\centering\arraybackslash}p{1.8cm} >{\centering\arraybackslash}p{1.8cm} >{\centering\arraybackslash}p{1.8cm} @{}}
\toprule
\multirow{2}{*}{\textbf{Method}} & \multicolumn{3}{c}{\textbf{AGD20K-Set 2 (Unseen)}} \\
\cmidrule(l){2-4}
& KLD $\downarrow$ & SIM $\uparrow$ & NSS $\uparrow$ \\
\midrule
Cross-View-AG+~\cite{luo2022learning} & 1.765 & 0.279 & 0.882 \\
LOCATE~\cite{li2023locate} & 1.405 & 0.372 & 1.157 \\
INTRA~\cite{jang2024intra} & 1.365 & 0.375 & 1.209 \\
WSMA~\cite{xu2024weakly} & 1.335 & 0.382 & 1.220 \\
LoopTrans~\cite{tang2025closed} & 1.247 & 0.403 & 1.315 \\
Moon et al.~\cite{moon2025selective} & 1.243 & 0.405 & \underline{1.368} \\
\midrule
\textbf{Ours -- InternVL3(2B)} & \underline{1.183} & \underline{0.429} & 1.323 \\
\textbf{Ours -- Qwen3-VL(2B)} & \textbf{1.060} & \textbf{0.455} & \textbf{1.514} \\
\bottomrule
\end{tabular}
\end{table}

\subsection{Comparison with WSAG-PLSP}

\begin{table}[t]
\caption{Performance comparison on AGD20K-Set 1 and Set 2. Best results are \textbf{bolded} and second-best results are \underline{underlined}.}
\centering
\renewcommand{\arraystretch}{1.1}
\begin{tabular}{l ccc ccc}
\toprule
\multirow{2}{*}{Method} & \multicolumn{3}{c}{AGD20K-Set 1} & \multicolumn{3}{c}{AGD20K-Set 2} \\
\cmidrule(lr){2-4} \cmidrule(lr){5-7}
& KLD$\downarrow$ & SIM$\uparrow$ & NSS$\uparrow$ & KLD$\downarrow$ & SIM$\uparrow$ & NSS$\uparrow$ \\
\midrule
WSAG-PLSP (ICLR'25) \cite{xu2025weaklysupervised}       & \textbf{0.890} & \textbf{0.510} & \textbf{1.547} & 1.153 & 0.437 & 1.418 \\
\midrule
\textbf{Ours} (2B)                         & 1.020 & \underline{0.459} & 1.406 & \underline{1.060} & \textbf{0.455} & \underline{1.514} \\
\textbf{Ours} (32B)                        & \underline{1.010} & 0.458 & \underline{1.424} & \textbf{1.043} & \textbf{0.455} & \textbf{1.549} \\
\bottomrule
\end{tabular}
\label{tab:performance_comparison}
\end{table}

While our main quantitative evaluation in Table 2 (main paper) focuses on comparable weakly supervised baselines in the zero-shot setting, we provide an additional comparison with the recently proposed WSAG-PLSP~\cite{xu2025weaklysupervised} in Table~\ref{tab:performance_comparison}. WSAG-PLSP introduces a paradigm that leverages Large Language Models (LLMs) to generate target texts, followed by a refinement process guided by ground-truth knowledge available in the AGD20K dataset. This access to dataset-specific priors during the training or refinement phase helps the model obtain more accurate textual and spatial alignment. Notably, WSAG-PLSP relies on dataset-specific refinement during training, whereas our method operates in a fully zero-shot manner.

As shown in Table~\ref{tab:performance_comparison}, this strategy is advantageous on the AGD20K-Set 1 (Seen) split, where WSAG-PLSP benefits from learning the target area distribution and therefore achieves higher performance. 
However, we observe that on the AGD20K-Set 2 (Unseen) split, our zero-shot framework demonstrates generalization, outperforming WSAG-PLSP across the evaluated metrics (e.g., a KLD of 1.043 vs. 1.153) without requiring task-specific data-driven refinement.

Given these methodological differences, quantitative metrics alone may not fully reflect the qualitative nuances and utility of the grounded affordances. To provide a broader evaluation, we conducted a blind user study  on AGD20K-Set 1 to compare the quality and human alignment of the affordance regions predicted by WSAG-PLSP and our proposed method, as detailed in ~\ref{sec:user_study}.

\section{User Study}
\label{sec:user_study}

\subsection{User Study Comparison with WSAG-PLSP}

\noindent \textbf{Study Setup}
To further evaluate the perceptual quality of the predicted affordance heatmaps, we conducted a user study comparing TokAG with three existing methods: LOCATE~\cite{li2023locate}, WSMA~\cite{xu2024weakly}, and WSAG-PLSP~\cite{xu2025weaklysupervised}.

We randomly sample 170 images from the AGD20K test set. For each image, we present the predicted affordance heatmaps produced by the four methods. The predictions from the four methods are displayed in random order and anonymized as \emph{Method~1--4} to ensure a blind comparison and avoid potential bias toward any specific method.

Participants are asked to select the predictions that best match the expected interaction for the given action--object pair. Multiple selections are allowed when several predictions are judged to equally capture this expected interaction. Furthermore, an option to select ``None'' is provided in cases where no prediction adequately reflects the expected interaction.

\noindent \textbf{Annotation Procedure}
Each participant evaluated the full set of 170 images independently. For each image, participants selected one or more predictions that best match the expected interaction for the given action--object pair. The annotations were recorded using a structured JSON format, where each entry corresponds to one image and stores the selected method indices. An example annotation entry is shown below:
\begin{verbatim}
{
  "image_id": "push_motorcycle_000123",
  "selected_methods": [2,4]
}
\end{verbatim}
This format allows multiple selections when several predictions are judged to equally match the expected interaction for the given action--object pair.

\noindent \textbf{Preference Rate}
Let $N$ denote the number of sampled images and $A$ denote the number of annotators.
For annotator $a$, let $S_i^{(a)}$ denote the set of methods selected for image $i$,
where multiple selections are allowed if several predictions are judged equally capture the expected interaction for the given action--object pair.
The selection rate of method $m$ for annotator $a$ is defined as

\begin{equation}
P_m^{(a)} = \frac{1}{N}\sum_{i=1}^{N}\mathbf{1}(m \in S_i^{(a)}),
\end{equation}
where $\mathbf{1}(m \in S_i^{(a)})$ equals 1 if method $m$ is selected for image $i$, and 0 otherwise.
We report the average preference rate across annotators:

\begin{equation}
\bar{P}_m = \frac{1}{A}\sum_{a=1}^{A} P_m^{(a)}.
\end{equation}

This metric measures the proportion of images for which a method is judged by annotators to be one of the predictions most consistent with the ground-truth affordance map. 
Because multiple selections are allowed, the preference rates of different methods do not necessarily sum to 100\%. 
% The result of our user study is shown in \cref{tab:user_study}. 

\begin{table}[t]
\centering
\caption{User study results on 170 randomly sampled AGD20K test images.
We report the preference rate (\%) for each method, where a higher value
indicates that the method is more frequently judged to best match the expected interaction for the given action--object pair.}
\label{tab:user_study}
\begin{tabular}{lccc}
\toprule
Method & Annotator 1 & Annotator 2 & Average \\
\midrule
LOCATE & 57.6 & 24.1 & 40.9 \\
WSMA & 54.7  & 21.8 & 38.3 \\
WSAG-PLSP & 56.5 & 25.3 & 40.9 \\
TokAG & \textbf{70.6}  & \textbf{35.9} & \textbf{53.3} \\
\bottomrule
\end{tabular}
\end{table}

% \noindent \textbf{Example Annotations}

\begin{figure}[htbp]
    \centering
    \begin{subfigure}{\linewidth}
        \centering
        \includegraphics[width=\linewidth]{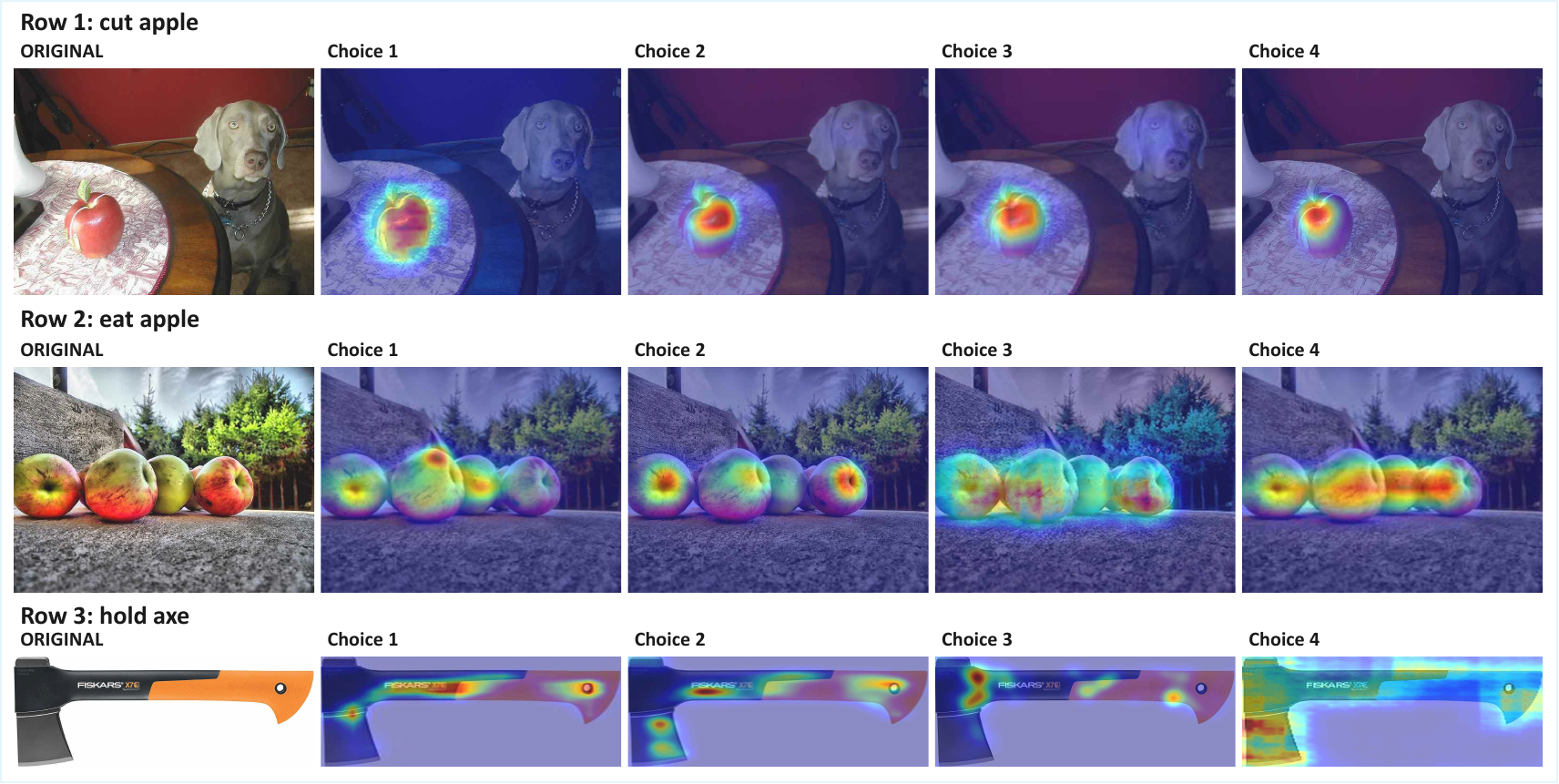}
        \label{fig:USER_STUDY_1}
    \end{subfigure}

    \centerline{\Large $\vdots$}
    \vspace{2em}

    \begin{subfigure}{\linewidth}
        \centering
        \includegraphics[width=\linewidth]{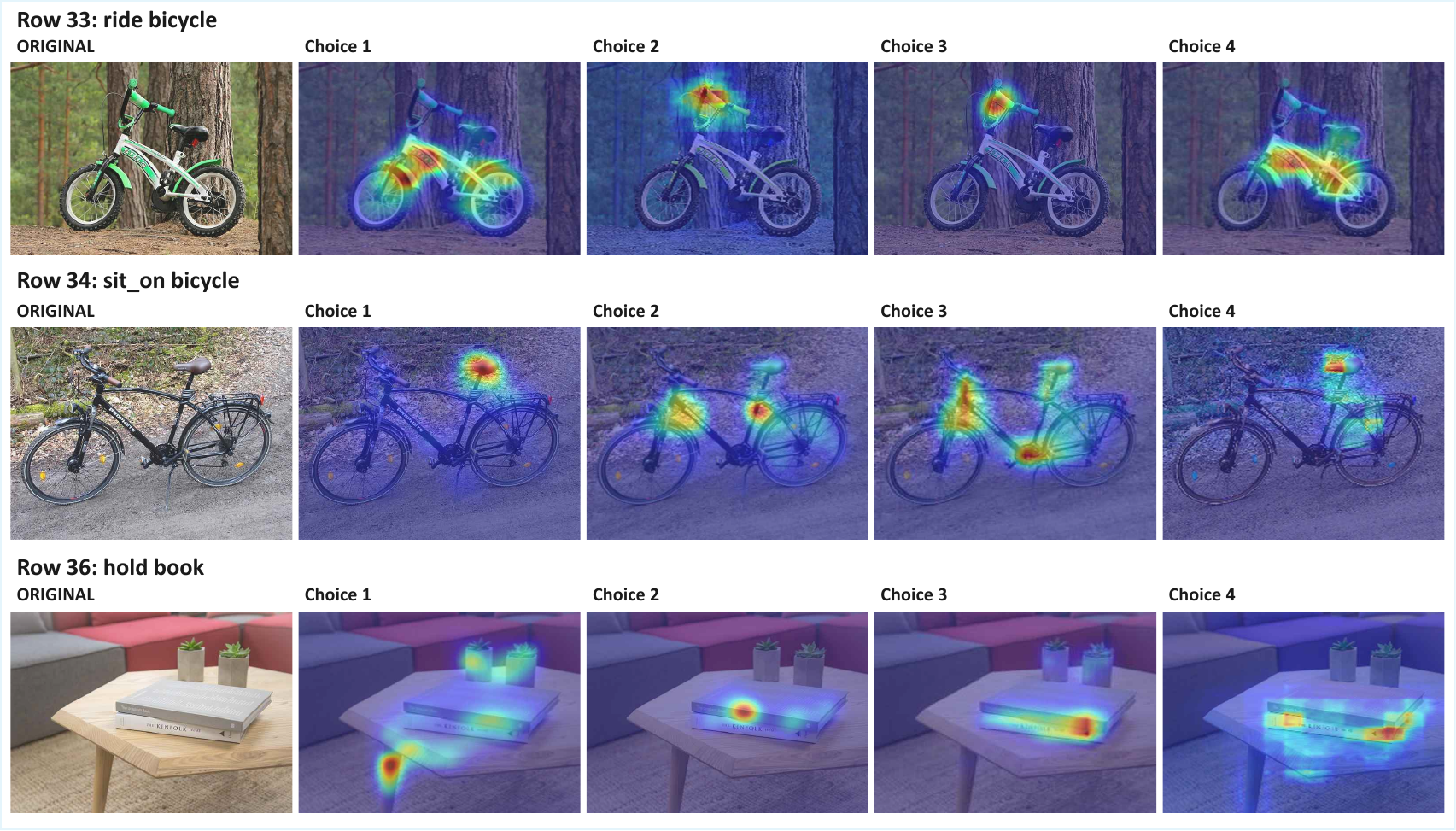}
        \label{fig:USER_STUDY_2}
    \end{subfigure}

     \centerline{\Large $\vdots$}
    
    \caption{
    Examples from the user study. For each row, the original input and the predicted affordance heatmaps from four methods were shown. 
    The order of the methods was randomized and anonymized as Choice 1–4 during the annotation process. 
    Participants were asked to select the prediction(s) that best match the expected interaction for the given action–object pair.
    }
    \label{fig:USER_STUDY_Combined}

\end{figure}

Table~\ref{tab:user_study} presents the quantitative results of the user study conducted on 170 sampled images. The selection rates indicate the frequency with which each method's prediction was judged by the annotators to be the closest to the expected interaction.
We include several example pages of the annotation interface as shown in Figure \ref{fig:USER_STUDY_Combined}. The full set of annotated images contains 170 samples used for the user study.

Across the evaluations from both annotators, our proposed TokAG records the highest average selection rate of 53.3\%. Specifically, TokAG was selected in 70.6\% and 35.9\% of the cases by Annotator 1 and Annotator 2, respectively. In comparison, the baseline methods yielded lower average rates: WSAG-PLSP and LOCATE both recorded 40.9\%, while WSMA recorded 38.3\%.

We observe a variance in the absolute selection rates between the two annotators. This difference reflects the inherent subjectivity of affordance evaluation and varying degrees of annotator leniency regarding multiple selections. However, despite this variance in absolute values, the relative performance trend among the evaluated methods remains consistent across both annotators. Overall, these quantitative outcomes indicate that the spatial heatmaps generated by our TokAG are more frequently evaluated by human annotators as best reflecting the expected interaction for the given action--object pair.

\section{Additional Qualitative Results}
\label{sec:qualitative_results}

\subsection{Additional Visual Comparisons}

\begin{figure}[p]
    \centering
    \includegraphics[width=\linewidth]{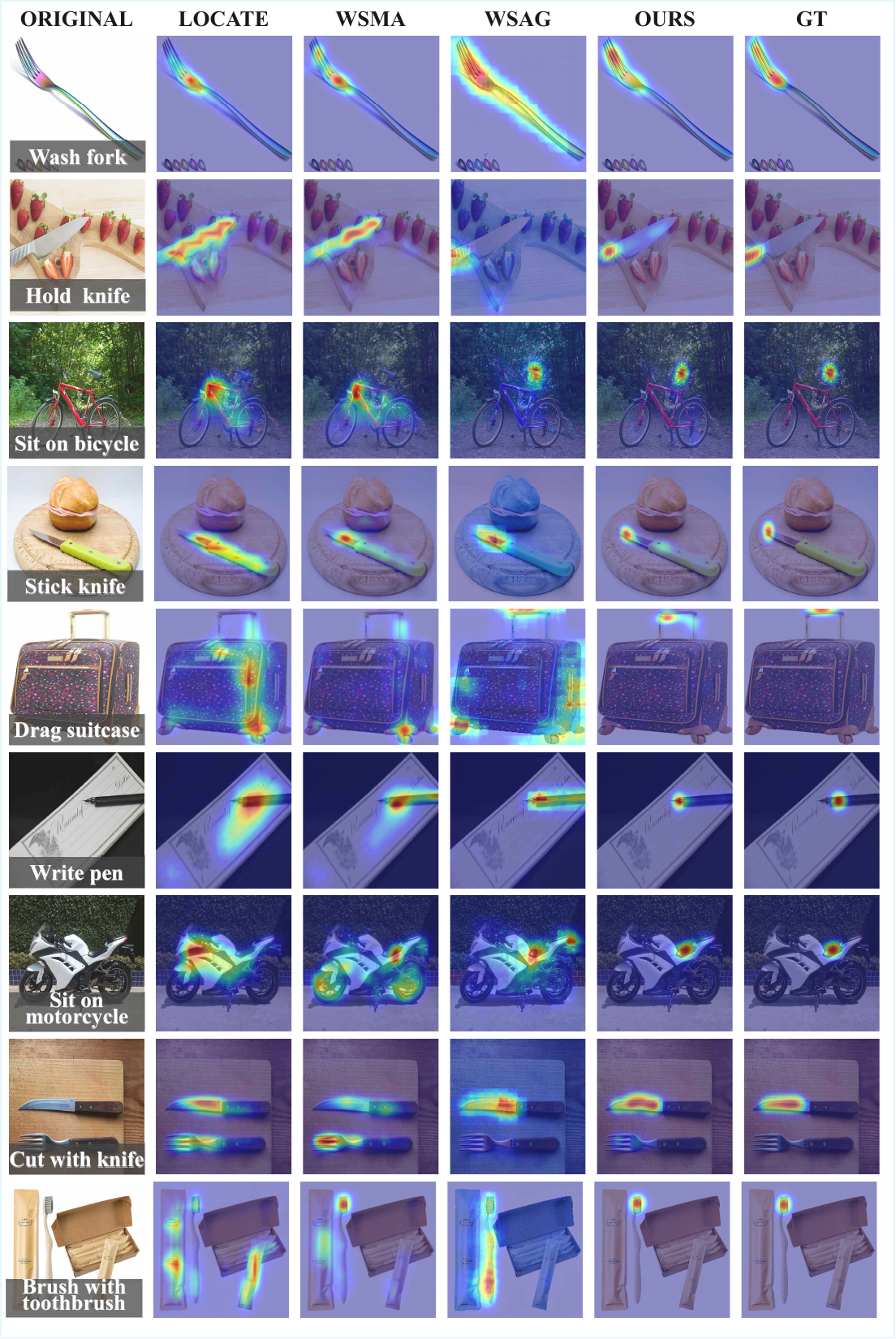}
    \caption{Qualitative comparison of affordance grounding results. We show predictions of TokAG (ours) together with LOCATE~\cite{li2023locate}, WSMA~\cite{xu2024weakly}, WSAG-PLSP~\cite{xu2025weaklysupervised}, and the ground truth (GT).}
    \label{fig:Qualitative_Results_1}
\end{figure}

\begin{figure}[p]
    \centering
    \includegraphics[width=\linewidth]{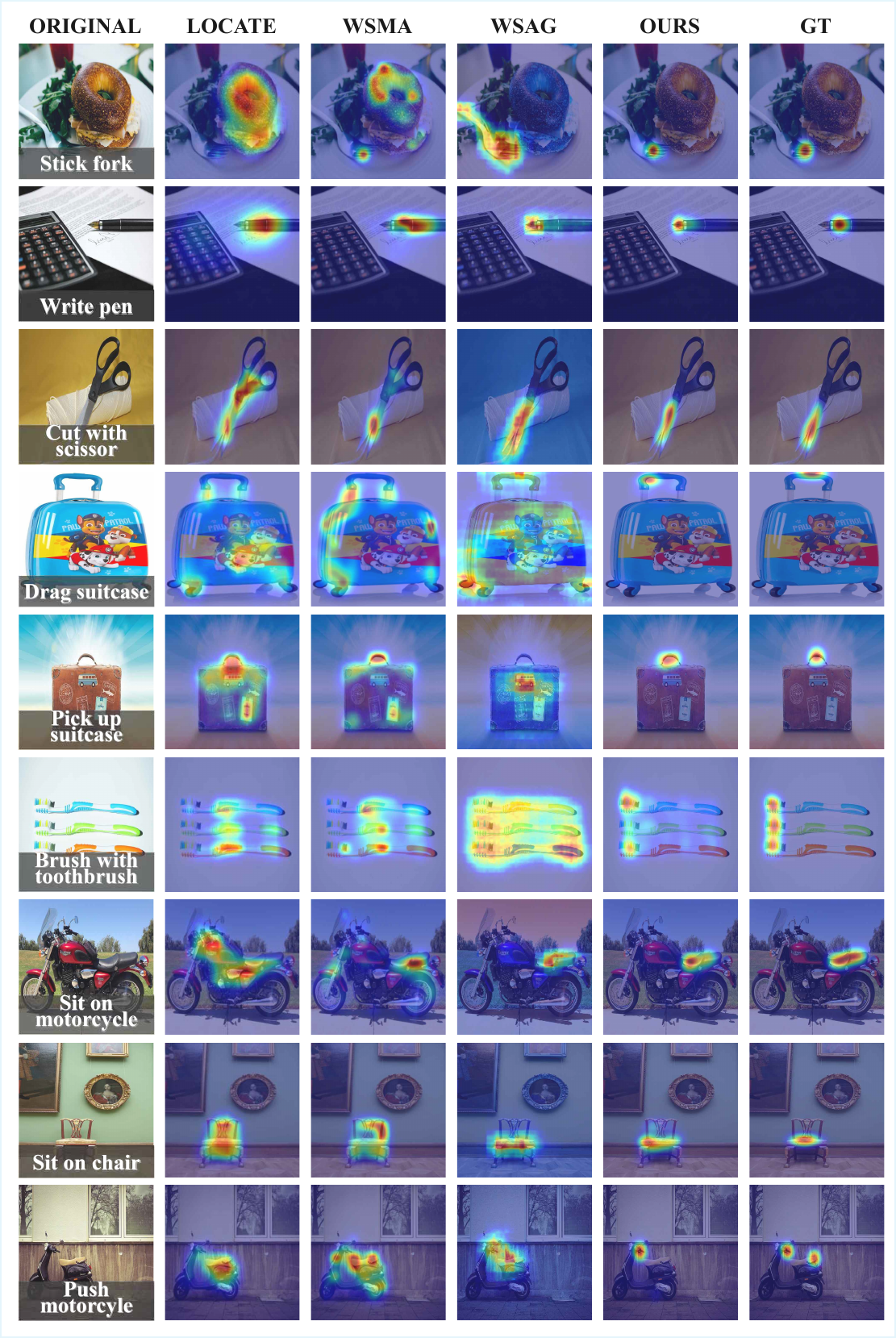}
    \caption{Qualitative comparison of affordance grounding results. We show predictions of TokAG (ours) together with LOCATE~\cite{li2023locate}, WSMA~\cite{xu2024weakly}, WSAG-PLSP~\cite{xu2025weaklysupervised}, and the ground truth (GT).}
    \label{fig:Qualitative_Results_2}
\end{figure}

As illustrated in Figure~\ref{fig:Qualitative_Results_1} and Figure~\ref{fig:Qualitative_Results_2}, we provide additional visualization results across various action--object categories. Across the evaluated categories, TokAG tends to produce more localized affordance predictions compared to existing approaches such as LOCATE~\cite{li2023locate}, WSMA~\cite{xu2024weakly}, and WSAG-PLSP~\cite{xu2025weaklysupervised}.

For instance, in action categories such as \textit{hold} and \textit{sit on}, our approach localizes the predicted heatmaps specifically to the expected interaction areas (e.g., handles and saddles). In contrast, the baseline methods often produce heatmaps that are more dispersed or mapped to non-interactive regions of the objects. 
These additional examples further show that TokAG generates affordance regions that align well with expected human interactions.

\subsection{Examples of Generated Sentences}

As described in our methodology, TokAG isolates target affordance regions through an output token selection mechanism based on the sentences generated by the LVLM (Large Vision-Language Model). To provide a clearer understanding of this process, Table~\ref{tab:generated_responses_examples} shows examples of the generated sentences together with the selected top tokens used for cross-attention extraction. These examples illustrate how the LVLM describes the functional parts associated with different action–object pairs and the selected token used to localize the affordance region in our method.

\small
\renewcommand{\arraystretch}{0.9}
\begin{xltabular}{\textwidth}{@{} l l l X @{}}
\caption{Examples of generated sentences for different action–object pairs. 
Given only the action and object categories, the LVLM generates a sentence describing the functional part involved in the interaction. 
The column ``Top Token'' shows the token selected by our token selection mechanism from the generated sentence.
}
\label{tab:generated_responses_examples} \\

\toprule
\textbf{Action} & \textbf{Object} & \textbf{Top Token} & \textbf{Generated Response} \\
\midrule
\endfirsthead

\multicolumn{4}{c}{{\tablename\ \thetable{} -- continued from previous page}} \\
\toprule
\textbf{Action} & \textbf{Object} & \textbf{Top Token} & \textbf{Generated Response} \\
\midrule
\endhead

\midrule
\multicolumn{4}{r}{{Continued on next page...}} \\
\endfoot

\bottomrule
\endlastfoot

% Real Data startt
\textit{carry} & \textit{skis} & \texttt{base} & The part of the skis used for `carry' is the base, which is the bottom surface of the ski that is in contact with the snow. \\
\textit{catch} & \textit{soccer ball} & \texttt{surface} & The part of the soccer ball used for `catch' is the surface that is in contact with the player's hand or glove, typically the top of the ball. \\
\textit{cut} & \textit{banana} & \texttt{peel} & The peel is used for cutting a banana. \\
\textit{cut\_with} & \textit{knife} & \texttt{blade} & The blade of the knife is used for `cut with'. \\
\textit{drink\_with} & \textit{cup} & \texttt{'s} & The part of the cup used for `drink with' is the cup's handle. \\
\textit{eat} & \textit{banana} & \texttt{yellow} & The part of the banana used for eating is the flesh, which is the yellow, sweet, and edible part of the fruit. \\
\textit{eat} & \textit{broccoli} & \texttt{ret} & The floret of the broccoli is used for eating. \\
\textit{hit} & \textit{axe} & \texttt{metal} & The part of the axe used for `hit' is the head, which is the metal part at the end of the axe. \\
\textit{hold} & \textit{axe} & \texttt{The} & The handle of the axe is used for holding. \\
\textit{hold} & \textit{cup} & \texttt{handle} & The handle of the cup is used for `hold'. \\
\textit{hold} & \textit{golf clubs} & \texttt{grip} & The part of the golf club used for `hold' is the grip. \\
\textit{hold} & \textit{knife} & \texttt{handle} & The handle of the knife is used for `hold'. \\
\textit{hold} & \textit{skis} & \texttt{base} & The part of the skis used for `hold' is the base, which is the bottom surface of the ski. \\
\textit{jump} & \textit{skis} & \texttt{base} & The part of the skis used for `jump' is the base, which is the bottom surface of the ski that provides the lift and propulsion for the jump. \\
\textit{kick} & \textit{soccer ball} & \texttt{ball} & The part of the soccer ball used for `kick' is the bottom. \\
\textit{lie\_on} & \textit{bed} & \texttt{mattress} & The part of the bed used for `lie on' is the mattress. \\
\textit{open} & \textit{refrigerator} & \texttt{top} & The top part of the refrigerator is used for `open'. \\
\textit{peel} & \textit{banana} & \texttt{outer} & The peel of the banana is the outer skin that is removed, typically from the stem end to the tip. \\
\textit{pick\_up} & \textit{skis} & \texttt{base} & The part of the skis used for `pick up' is the base, which is the bottom surface of the ski. \\
\textit{pour} & \textit{cup} & \texttt{sp} & The part of the cup used for pouring is the spout. \\
\textit{push} & \textit{bicycle} & \texttt{the} & The part of the bicycle used for `push' is the pedal. \\
\textit{ride} & \textit{bicycle} & \texttt{frame} & The part of the bicycle used for `ride' is the frame, which is the main structure that supports the rider and the wheels. \\
\textit{sip} & \textit{cup} & \texttt{rim} & The part of the cup used for `sip' is the rim. \\
\textit{sit\_on} & \textit{bed} & \texttt{head} & The headboard is used for `sit on' with the bed. \\
\textit{sit\_on} & \textit{bicycle} & \texttt{seat} & The seat of the bicycle is used for `sit on'. \\
\textit{stick} & \textit{knife} & \texttt{handle} & The handle of the knife is used for `stick'. \\
\textit{swing} & \textit{golf clubs} & \texttt{head} & The part of the golf club used for `swing' is the club head. \\
\textit{take\_photo} & \textit{camera} & \texttt{lens} & The lens is used for `take photo' with the camera. \\
\textit{throw} & \textit{basketball} & \texttt{surface} & The part of the basketball used for `throw' is the ball itself, specifically the surface that is in contact with the hand or arm during the throwing motion. \\
\textit{type\_on} & \textit{laptop} & \texttt{keyboard} & The keyboard is the part of the laptop used for `type on'. \\
\textit{wash} & \textit{cup} & \texttt{the} & The part of the cup used for `wash' is the inner surface of the cup, specifically the inside of the cup, which is cleaned with a wash cloth or sponge. \\
\textit{wash} & \textit{knife} & \texttt{blade} & The blade is used for washing. \\

\end{xltabular}

\subsection{Quantitative and Qualitative Results on an Additional Dataset: EPIC-Aff}

\begin{figure}[p]
\begin{center}
    \centering

    % =========================================================
    % --- Row 1: cut carrot ---
    % =========================================================
    % Add action label
    \begin{minipage}[c]{0.18\linewidth}
        \centerline{\scriptsize{   Action: cut carrot}}
    \end{minipage}\hfill
    \begin{minipage}[c]{0.18\linewidth}
        \centerline{}
    \end{minipage}\hfill
    \begin{minipage}[c]{0.18\linewidth}
        \centerline{}
    \end{minipage}\hfill    
    \begin{minipage}[c]{0.18\linewidth}
        \centerline{}
    \end{minipage}\hfill
    \begin{minipage}[c]{0.18\linewidth}
        \centerline{}
    \end{minipage}

    % Images
    \begin{minipage}[c]{0.18\linewidth}
        \includegraphics[width=\linewidth,height=\linewidth,keepaspectratio=false]{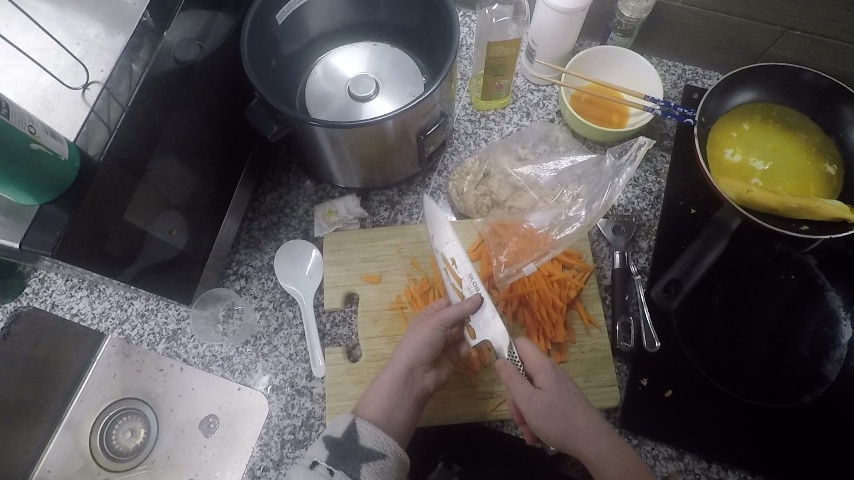}
    \end{minipage}\hfill
    \begin{minipage}[c]{0.18\linewidth}
        \includegraphics[width=\linewidth,height=\linewidth,keepaspectratio=false]{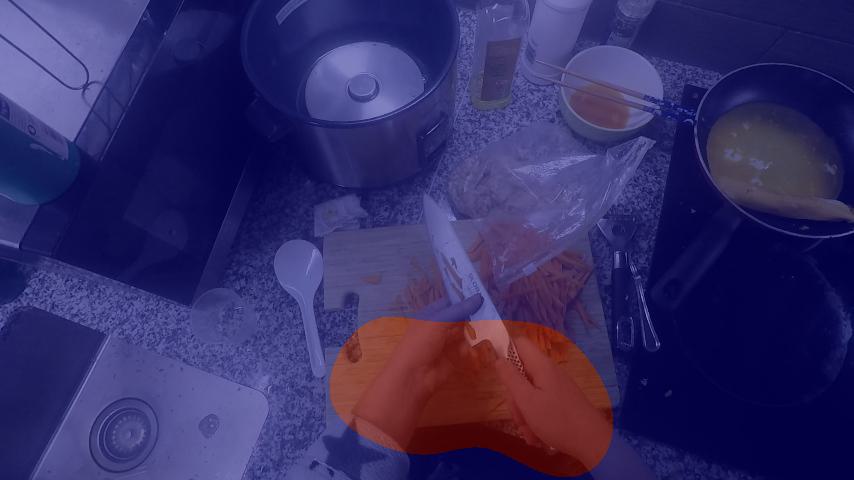}
    \end{minipage}\hfill
    \begin{minipage}[c]{0.18\linewidth}
        \includegraphics[width=\linewidth,height=\linewidth,keepaspectratio=false]{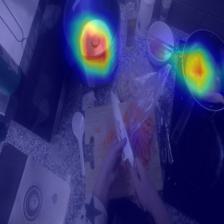}
    \end{minipage}\hfill
    \begin{minipage}[c]{0.18\linewidth}
        \includegraphics[width=\linewidth,height=\linewidth,keepaspectratio=false]{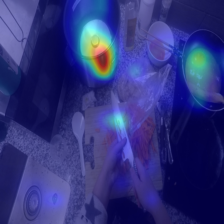}
    \end{minipage}\hfill    
    \begin{minipage}[c]{0.18\linewidth}
        \includegraphics[width=\linewidth,height=\linewidth,keepaspectratio=false]{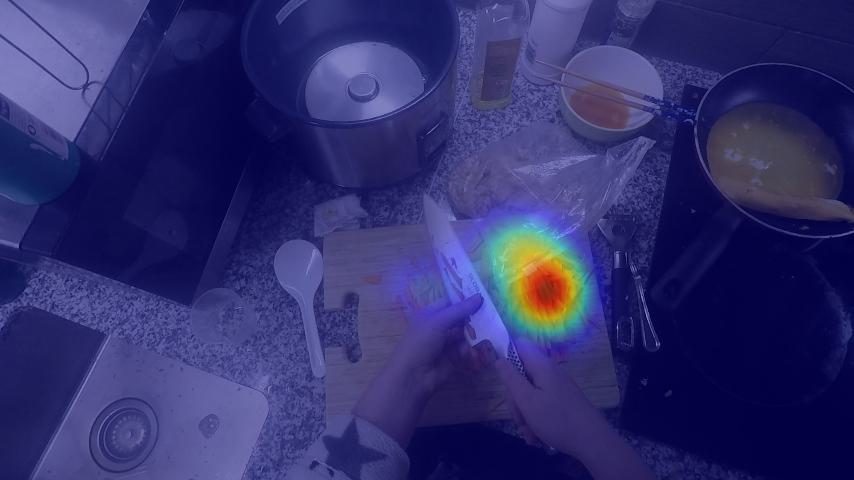}
    \end{minipage}

    \vspace{6pt}

    % =========================================================
    % --- Row 2: open bottle ---
    % =========================================================
    % Add action label
    \begin{minipage}[c]{0.18\linewidth}
        \centerline{\scriptsize{   Action: open bottle}}
    \end{minipage}\hfill
    \begin{minipage}[c]{0.18\linewidth}
        \centerline{}
    \end{minipage}\hfill
    \begin{minipage}[c]{0.18\linewidth}
        \centerline{}
    \end{minipage}\hfill    
    \begin{minipage}[c]{0.18\linewidth}
        \centerline{}
    \end{minipage}\hfill
    \begin{minipage}[c]{0.18\linewidth}
        \centerline{}
    \end{minipage}

    % Images 
    \begin{minipage}[c]{0.18\linewidth}
        \includegraphics[width=\linewidth,height=\linewidth,keepaspectratio=false]{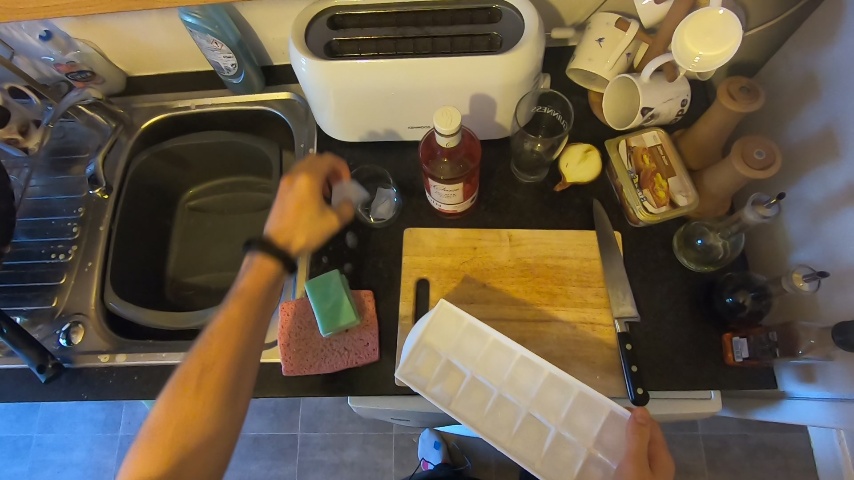}
    \end{minipage}\hfill
    \begin{minipage}[c]{0.18\linewidth}
        \includegraphics[width=\linewidth,height=\linewidth,keepaspectratio=false]{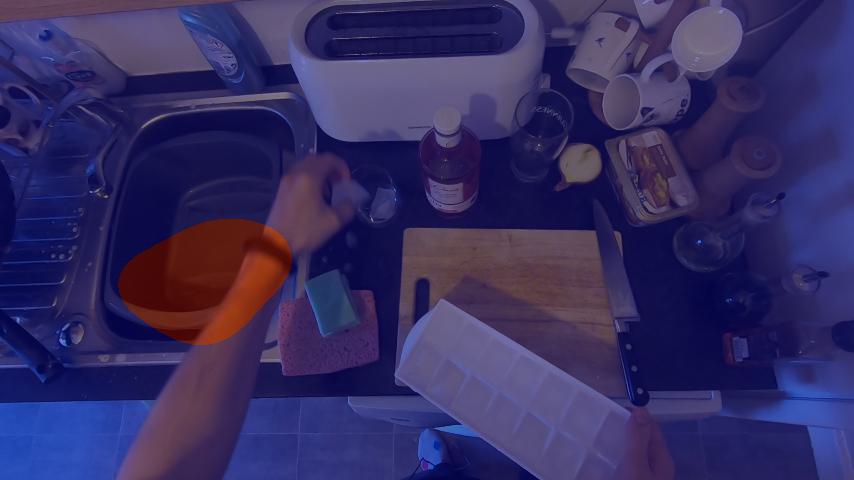}
    \end{minipage}\hfill
    \begin{minipage}[c]{0.18\linewidth}
        \includegraphics[width=\linewidth,height=\linewidth,keepaspectratio=false]{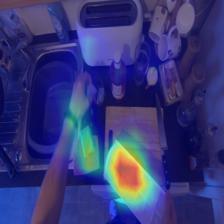}
    \end{minipage}\hfill
    \begin{minipage}[c]{0.18\linewidth}
        \includegraphics[width=\linewidth,height=\linewidth,keepaspectratio=false]{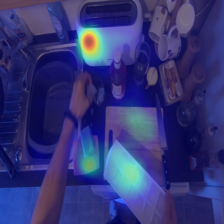}
    \end{minipage}\hfill
    \begin{minipage}[c]{0.18\linewidth}
        \includegraphics[width=\linewidth,height=\linewidth,keepaspectratio=false]{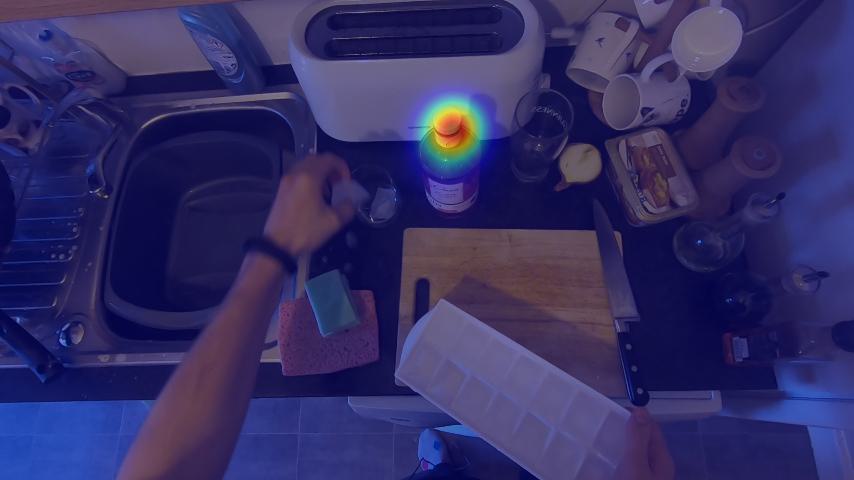}
    \end{minipage}
    
    \vspace{6pt}

    % =========================================================
    % --- Row 3: open oven ---
    % =========================================================
    % Add action label
    \begin{minipage}[c]{0.18\linewidth}
        \centerline{\scriptsize{   Action: open oven}}
    \end{minipage}\hfill
    \begin{minipage}[c]{0.18\linewidth}
        \centerline{}
    \end{minipage}\hfill
    \begin{minipage}[c]{0.18\linewidth}
        \centerline{}
    \end{minipage}\hfill    
    \begin{minipage}[c]{0.18\linewidth}
        \centerline{}
    \end{minipage}\hfill
    \begin{minipage}[c]{0.18\linewidth}
        \centerline{}
    \end{minipage}
    
    % Images 
    \begin{minipage}[c]{0.18\linewidth}
        \includegraphics[width=\linewidth,height=\linewidth,keepaspectratio=false]{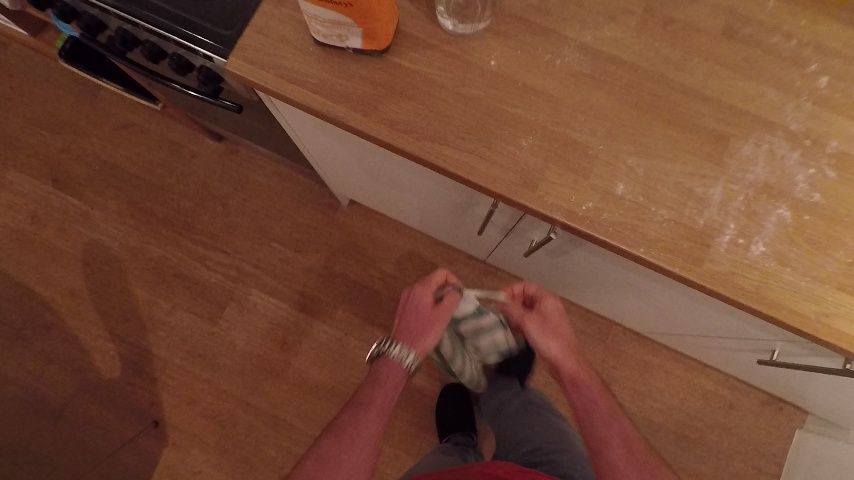}
    \end{minipage}\hfill
    \begin{minipage}[c]{0.18\linewidth}
        \includegraphics[width=\linewidth,height=\linewidth,keepaspectratio=false]{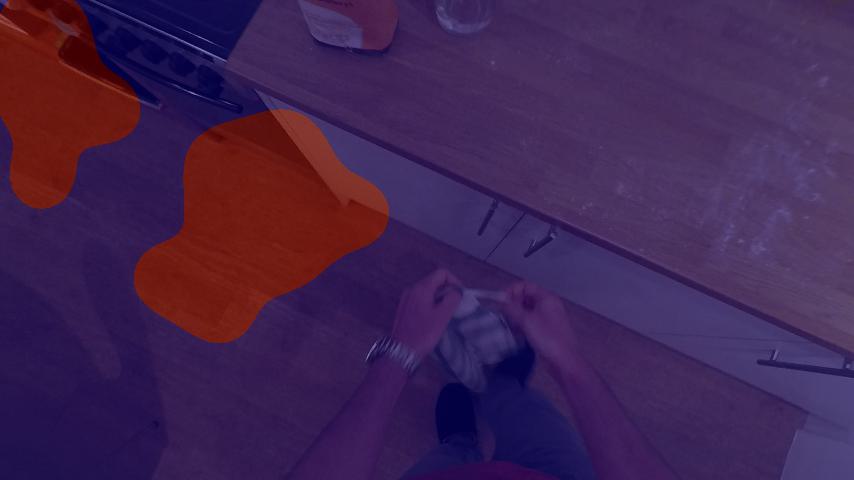}
    \end{minipage}\hfill
    \begin{minipage}[c]{0.18\linewidth}
        \includegraphics[width=\linewidth,height=\linewidth,keepaspectratio=false]{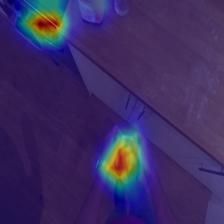}
    \end{minipage}\hfill
    \begin{minipage}[c]{0.18\linewidth}
        \includegraphics[width=\linewidth,height=\linewidth,keepaspectratio=false]{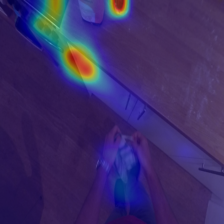}
    \end{minipage}\hfill
    \begin{minipage}[c]{0.18\linewidth}
        \includegraphics[width=\linewidth,height=\linewidth,keepaspectratio=false]{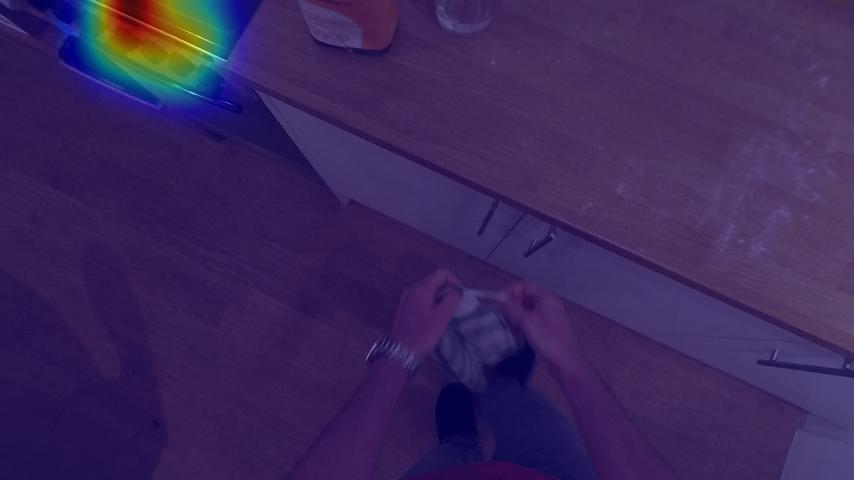}
    \end{minipage}

    \vspace{6pt}

    % =========================================================
    % --- Row 4: wash bowl ---
    % =========================================================
    % Add action label
    \begin{minipage}[c]{0.18\linewidth}
        \centerline{\scriptsize{Action: wash bowl}}
    \end{minipage}\hfill
    \begin{minipage}[c]{0.18\linewidth}
        \centerline{}
    \end{minipage}\hfill
    \begin{minipage}[c]{0.18\linewidth}
        \centerline{}
    \end{minipage}\hfill    
    \begin{minipage}[c]{0.18\linewidth}
        \centerline{}
    \end{minipage}\hfill
    \begin{minipage}[c]{0.18\linewidth}
        \centerline{}
    \end{minipage}
    
    % Images 
    \begin{minipage}[c]{0.18\linewidth}
        \includegraphics[width=\linewidth,height=\linewidth,keepaspectratio=false]{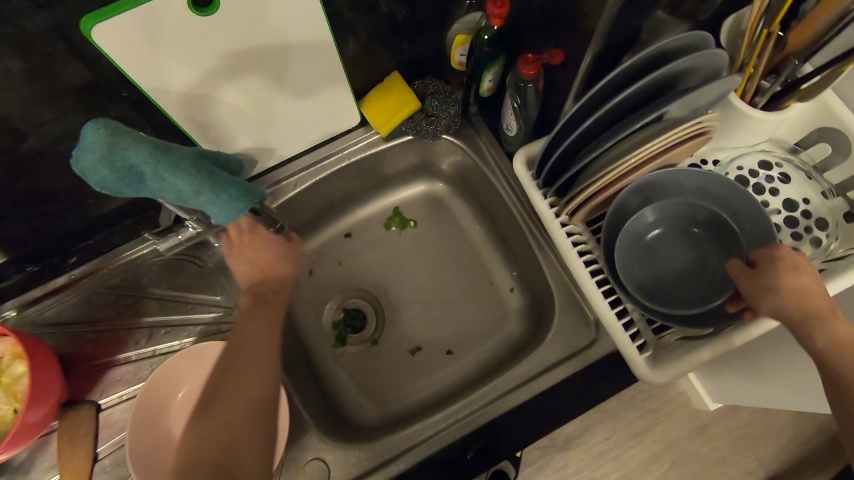}
    \end{minipage}\hfill
    \begin{minipage}[c]{0.18\linewidth}
        \includegraphics[width=\linewidth,height=\linewidth,keepaspectratio=false]{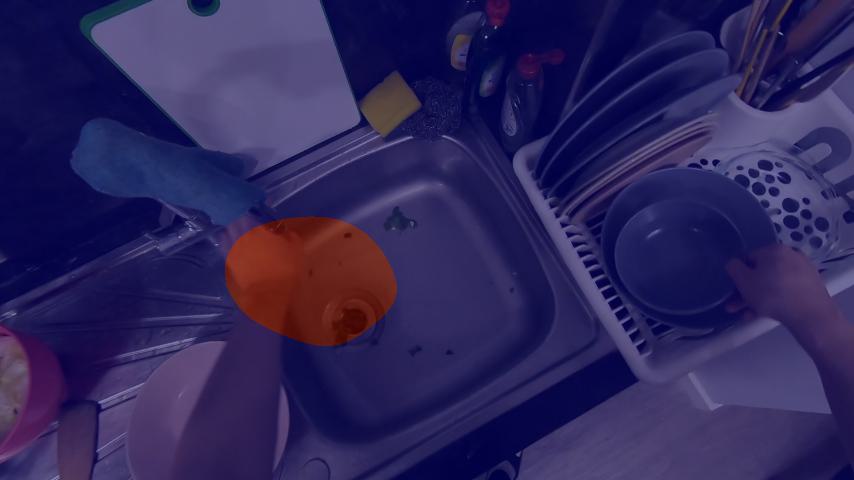}
    \end{minipage}\hfill
    \begin{minipage}[c]{0.18\linewidth}
        \includegraphics[width=\linewidth,height=\linewidth,keepaspectratio=false]{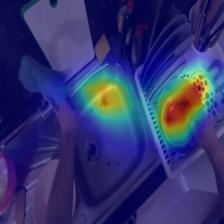}
    \end{minipage}\hfill
    \begin{minipage}[c]{0.18\linewidth}
        \includegraphics[width=\linewidth,height=\linewidth,keepaspectratio=false]{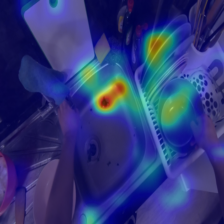}
    \end{minipage}\hfill
    \begin{minipage}[c]{0.18\linewidth}
        \includegraphics[width=\linewidth,height=\linewidth,keepaspectratio=false]{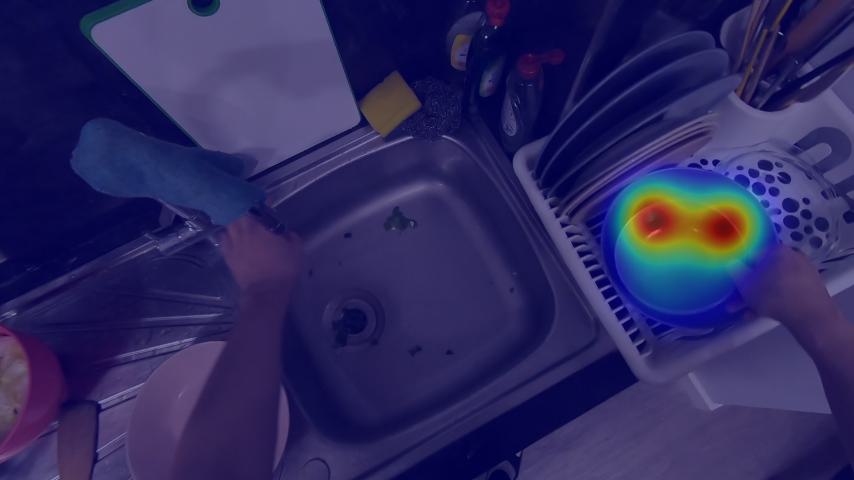}
    \end{minipage}

    \vspace{6pt}

    % =========================================================
    % --- Row 5: wash fork ---
    % =========================================================
    % Add action label
    \begin{minipage}[c]{0.18\linewidth}
        \centerline{\scriptsize{Action: wash fork}}
    \end{minipage}\hfill
    \begin{minipage}[c]{0.18\linewidth}
        \centerline{}
    \end{minipage}\hfill
    \begin{minipage}[c]{0.18\linewidth}
        \centerline{}
    \end{minipage}\hfill    
    \begin{minipage}[c]{0.18\linewidth}
        \centerline{}
    \end{minipage}\hfill
    \begin{minipage}[c]{0.18\linewidth}
        \centerline{}
    \end{minipage}
    
    % Images 
    \begin{minipage}[c]{0.18\linewidth}
        \includegraphics[width=\linewidth,height=\linewidth,keepaspectratio=false]{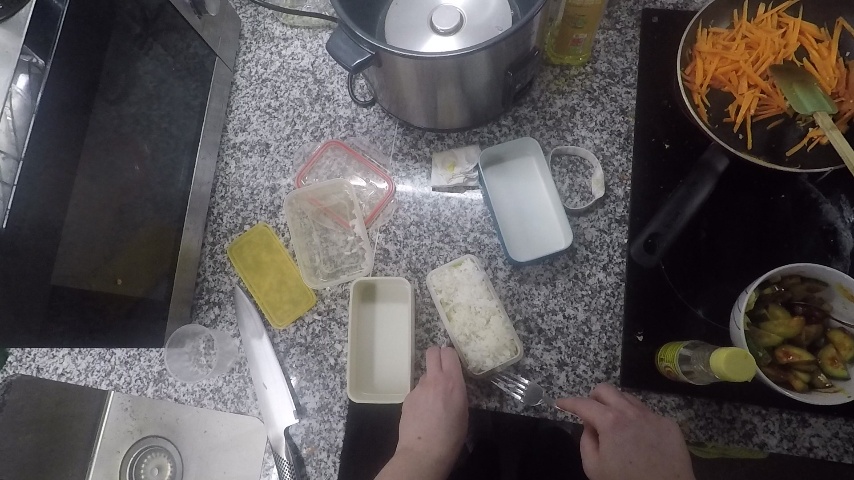}
    \end{minipage}\hfill
    \begin{minipage}[c]{0.18\linewidth}
        \includegraphics[width=\linewidth,height=\linewidth,keepaspectratio=false]{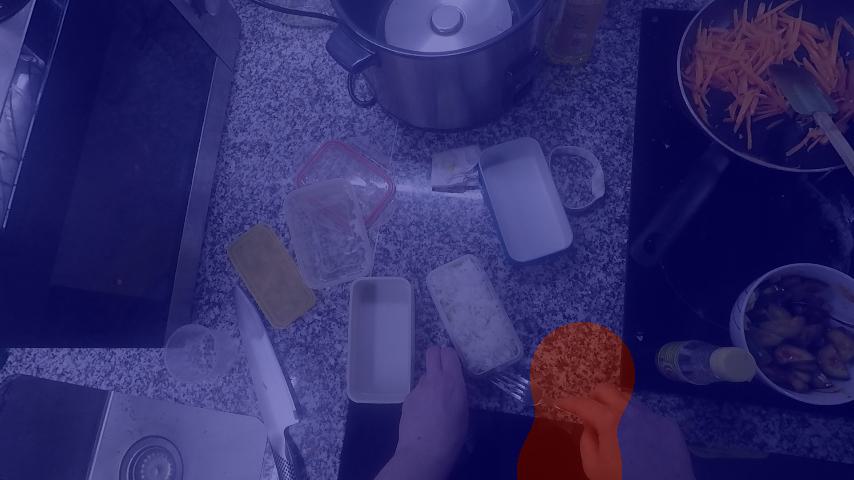}
    \end{minipage}\hfill
    \begin{minipage}[c]{0.18\linewidth}
        \includegraphics[width=\linewidth,height=\linewidth,keepaspectratio=false]{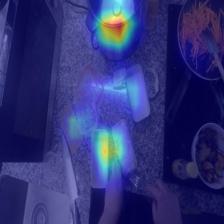}
    \end{minipage}\hfill
    \begin{minipage}[c]{0.18\linewidth}
        \includegraphics[width=\linewidth,height=\linewidth,keepaspectratio=false]{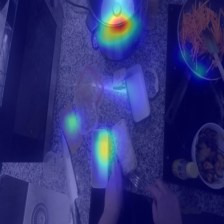}
    \end{minipage}\hfill
    \begin{minipage}[c]{0.18\linewidth}
        \includegraphics[width=\linewidth,height=\linewidth,keepaspectratio=false]{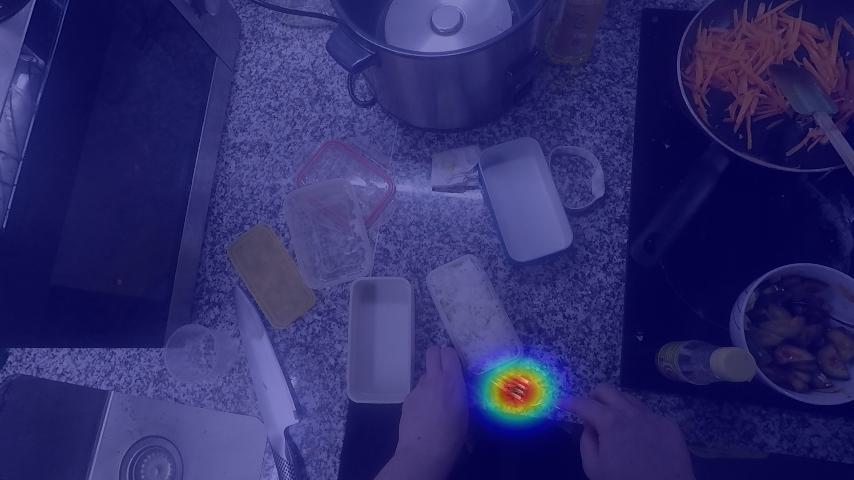}
    \end{minipage}

    \vspace{6pt}

    % =========================================================
    % --- Row 6: wash knife ---
    % =========================================================
    % Add action label
    \begin{minipage}[c]{0.18\linewidth}
        \centerline{\scriptsize{Action: wash knife}}
    \end{minipage}\hfill
    \begin{minipage}[c]{0.18\linewidth}
        \centerline{}
    \end{minipage}\hfill
    \begin{minipage}[c]{0.18\linewidth}
        \centerline{}
    \end{minipage}\hfill    
    \begin{minipage}[c]{0.18\linewidth}
        \centerline{}
    \end{minipage}\hfill
    \begin{minipage}[c]{0.18\linewidth}
        \centerline{}
    \end{minipage}
    
    % Images 
    \begin{minipage}[c]{0.18\linewidth}
        \includegraphics[width=\linewidth,height=\linewidth,keepaspectratio=false]{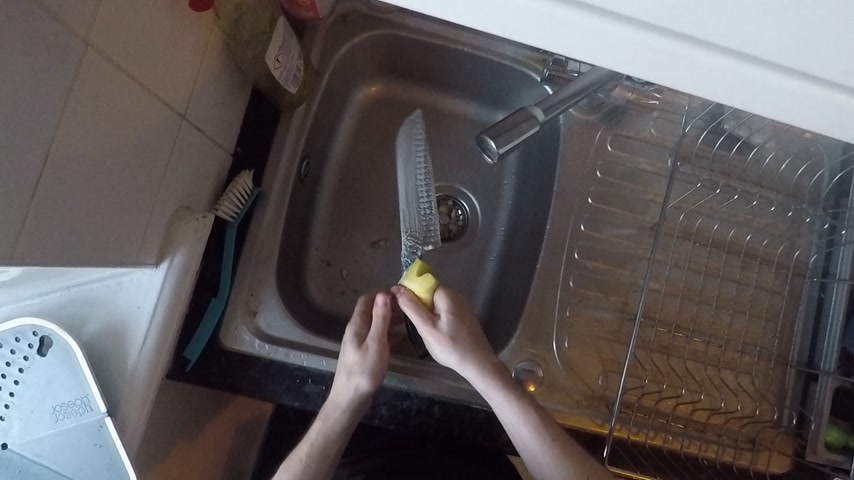}
    \end{minipage}\hfill
    \begin{minipage}[c]{0.18\linewidth}
        \includegraphics[width=\linewidth,height=\linewidth,keepaspectratio=false]{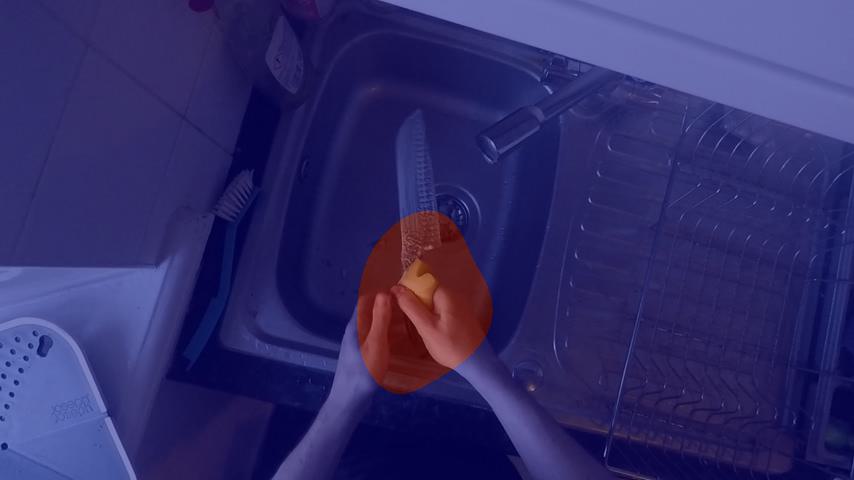}
    \end{minipage}\hfill
    \begin{minipage}[c]{0.18\linewidth}
        \includegraphics[width=\linewidth,height=\linewidth,keepaspectratio=false]{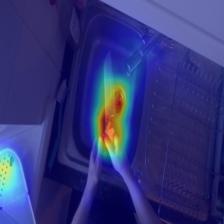}
    \end{minipage}\hfill
    \begin{minipage}[c]{0.18\linewidth}
        \includegraphics[width=\linewidth,height=\linewidth,keepaspectratio=false]{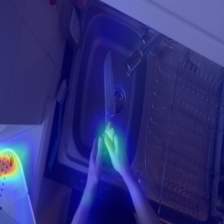}
    \end{minipage}\hfill
    \begin{minipage}[c]{0.18\linewidth}
        \includegraphics[width=\linewidth,height=\linewidth,keepaspectratio=false]{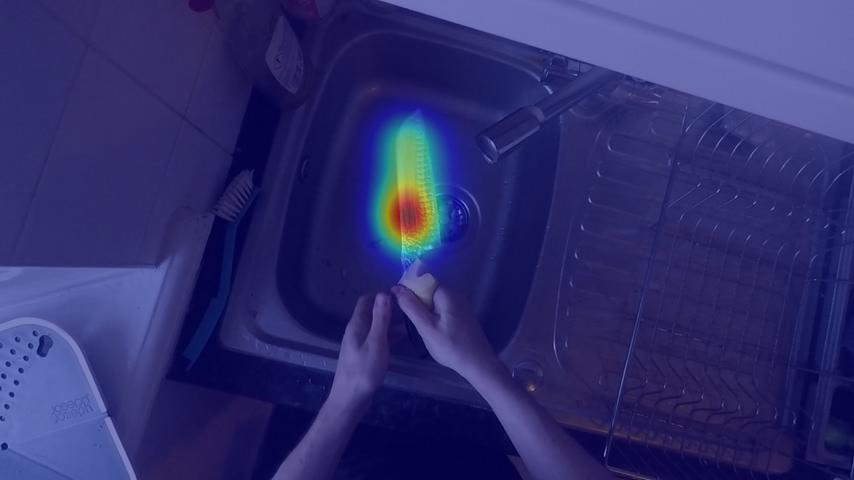}
    \end{minipage}

    \vspace{4pt}
    % =========================================================
    % --- Bottom Column Labels ---
    % =========================================================
    \begin{minipage}[c]{0.18\linewidth}
        \centerline{\scriptsize{Input Image}}
    \end{minipage}\hfill
    \begin{minipage}[c]{0.18\linewidth}
        \centerline{\scriptsize{GT}}
    \end{minipage}\hfill
    \begin{minipage}[c]{0.18\linewidth}
        \centerline{\scriptsize{LOCATE}}
    \end{minipage}\hfill    
    \begin{minipage}[c]{0.18\linewidth}
        \centerline{\scriptsize{WSMA}}
    \end{minipage}\hfill
    \begin{minipage}[c]{0.18\linewidth}
        \centerline{\scriptsize{Ours}}
    \end{minipage}

\vspace{-0.2em}
\captionof{figure}{Qualitative comparison on the EPIC-Aff dataset.}
\label{fig:Qualitative analysis on EPIC-Aff}
\end{center}
\end{figure}

To further evaluate the proposed TokAG, we conduct additional experiment on the EPIC-Aff dataset~\cite{mur2023multi}, a benchmark derived from the first-person perspective video dataset EPIC-Kitchens~\cite{damen2020epic}. To match the scale of the AGD20K test set (540 images), we randomly sampled 500 images from the unseen split condition of EPIC-Aff. As summarized in Table~\ref{tab:epic_comparison}, TokAG shows improved results compared to existing SOTA methods. In addition, we observe that the numbers of different evaluation metrics for all models on EPIC-Aff are lower than those on AGD20K; this discrepancy may come from the dataset characteristics of EPIC-Aff, which contains fewer object-centric ground-truth regions. Despite these characteristics, TokAG isolates the target affordance areas, as shown in Figure~\ref{fig:Qualitative analysis on EPIC-Aff}. TokAG localizes the functional regions, indicating that the proposed token-based grounding framework transfers to different domain distributions without further fine-tuning.

% --- [수정] 1. 독립된 테이블 환경 (Quantitative Results) ---
\begin{table}[htp]
    \centering
    \caption{Quantitative comparison on the EPIC-Aff dataset.}
    \label{tab:epic_comparison}
    \small
    \setlength{\tabcolsep}{18pt}
    \begin{tabular}{lccc}
    \toprule
    \textbf{Method} & \textbf{KLD} $\downarrow$ & \textbf{SIM} $\uparrow$ & \textbf{NSS} $\uparrow$ \\ 
    \midrule
    LOCATE          & 3.901          & 0.071          & 0.366          \\
    WSMA            & 3.372          & 0.062          & 0.413          \\ 
    \textbf{TokAG (Ours)} & \textbf{3.367} & \textbf{0.122} & \textbf{0.927} \\ 
    \bottomrule
    \end{tabular}
\end{table}

